\documentclass{article}

\PassOptionsToPackage{numbers}{natbib}
\usepackage[preprint]{neurips_2022}

\usepackage[utf8]{inputenc} % allow utf-8 input
\usepackage[T1]{fontenc}    % use 8-bit T1 fonts
\usepackage{hyperref}       % hyperlinks
\usepackage{url}            % simple URL typesetting
\usepackage{booktabs}       % professional-quality tables
\usepackage{amsfonts}       % blackboard math symbols
\usepackage{nicefrac}       % compact symbols for 1/2, etc.
\usepackage{microtype}      % microtypography
\usepackage{xcolor}         % colors

\usepackage{graphicx}
\usepackage{subcaption}
\usepackage{amssymb}
\usepackage{mathtools}
\usepackage{amsthm}

\usepackage{float}
\usepackage{placeins}
\usepackage{algorithmic}
\usepackage{algorithm}

\usepackage{changepage}

\title{Revisiting Self-Training \texorpdfstring{\\}{}
             with Regularized Pseudo-Labeling for Tabular Data}

\author{%
  Minwook Kim$^{1,2}$ \quad Juseong Kim$^{1,2}$ \quad Giltae Song$^{1,2}$ \\
  $^1$School of Computer Science and Engineering, Pusan National University, Busan, Republic of Korea \\
  $^2$ Center for Artificial Intelligence Research, Pusan National University, Busan, Republic of Korea \\
  \texttt{\{kmiiiaa,kjseong,gsong\}@pusan.ac.kr}\\
}

\begin{document}

\maketitle

\begin{abstract}
  Recent progress in semi- and self-supervised learning has caused a rift in the long-held belief about the need for an enormous amount of labeled data for machine learning and the irrelevancy of unlabeled data.
  Although it has been successful in various data, there is no dominant semi- and self-supervised learning method that can be generalized for tabular data (i.e. most of the existing methods require appropriate tabular datasets and architectures).
  In this paper, we revisit self-training which can be applied to any kind of algorithm including the most widely used architecture, gradient boosting decision tree, and introduce curriculum pseudo-labeling (a state-of-the-art pseudo-labeling technique in image) for a tabular domain.
  Furthermore, existing pseudo-labeling techniques do not assure the cluster assumption when computing confidence scores of pseudo-labels generated from unlabeled data. 
  To overcome this issue, we propose a novel pseudo-labeling approach that regularizes the confidence scores based on the likelihoods of the pseudo-labels so that more reliable pseudo-labels which lie in high density regions can be obtained. 
  We exhaustively validate the superiority of our approaches using various models and tabular datasets.
\end{abstract}

\section{Introduction}\label{sec:Introduction}

Supervised learning has had great success that outperforms human beings or heuristic algorithms in various tasks \cite{machine_learning, deeplearning}. 
These achievements, however, essentially rely on large labeled data.
Since it requires significant financial costs and human resources to label a collection of data \cite{semantic_segmentation, intro_weakly_sl, toward_weakly_sl}, a large amount of data still remain unlabeled. 
This limits to training a model with supervised learning because we are not able to annotate all the unlabeled data.
In the fields where data can be easily obtained but be difficult to be labeled (e.g.\ computer vision or natural language processing), semi- and self-supervised learning are one of the promising learning paradigms to boost the performance of its model by leveraging unlabeled data. 
However, unlike computer vision and natural language processing, semi- and self-supervised learning have not been quite successful for tabular data yet.

Tabular data is composed of rows to represent instances, and columns features and labels. 
Hence, obtaining features of each instance is similar  to its label. 
Since the process of collecting features usually include  labeling, there are not many cases where data remain unlabeled. 
Moreover, in contrast to computer vision or natural language processing fields where people can use other datasets via transfer learning for different tasks with the pretrained model, transfer learning for domains where tabular data is used is quite tricky \cite{survey_of_transfer_learning, survey_tabular_nn, transfer_tabular}.
Most images or text are homogeneous; they have common structures. 
However, most tabular datasets are heterogeneous.
While all images, for example, consist of 3-color channels or gray-scale, and text data share common grammar and words, most tabular datasets have a different feature set for their purpose. 
This causes difficulty in applying transfer learning to tabular datasets for other tasks than its original purpose.
This heterogeneous nature among tabular datasets is why semi- and self-supervised learning has not been much attractive to the tabular data field yet.

Interestingly, there exist some tabular data that their labeling process is not entirely the same as collecting their features. 
Especially this is quite common in biomedicine and healthcare domains such as electronic healthcare records which are representative data in mainly tabular format \cite{transfer_tabular}.
For instance, tasks such as predicting the prognosis of patients, which is one of the important questions need clinical experts to continue follow-ups with the patients to label the data \cite{mimic3, mimic4, kamir}.
These continuous follow-ups often fail due to various reasons caused by the patients as well as the experts, which result in generating unlabeled data.
Also, labeling data in healthcare sometimes requires expensive additional examinations in hospitals such as Magnetic Resonance Imaging (MRI) or Computed Tomography Scan (CT-Scan) \cite{dicom-rt, reverse_remodeling}.
In this case, data often remain unlabeled inevitably since additional examinations are not able to be conducted due to budget limits.
This causes vast amounts of unlabeled data in biomedicine and healthcare. 

Therefore, there have been some attempts to apply semi- and self-supervised learning to utilize unlabeled data during the training process for these tabular data \cite{vime, contrastive_misup, transfer_tabular}.
However, most semi- and self-supervised learning methods for the tabular data have a strong dependency on the structure of their models with special objectives such as reconstruction loss and contrastive loss, and the dataset composition such as feature types and imbalance ratio per label. 
This is why they have not been quite successful for tabular data.   
In addition, according to Kaggle, while more than 50\% of researchers have used gradient boosting decision trees in 2021 \cite{kaggle_2021_state}, the decision tree based methods are incompatible with the gradient descent step that is required for applying existing semi- and self-supervised methods.

In this study, we revisit self-training, which leverages unlabeled data by generating pseudo-labels and is a universally applicable semi-supervised  method regardless of their model and dataset composition.
As pseudo-labeling methods have been out of date for tabular data \cite{tabtransformer, scarf}, we introduce a state-of-the-art pseudo-labeling method called curriculum pseudo-labeling that has been used in the computer vision field where has shown continuous progress for self-training \cite{curriculum_labeling}.

We also address another issue with regard to pseudo-labeling which uses prediction confidence scores (or prediction probabilities) to obtain pseudo-labels \cite{pseudo_label, pseudo_label5}.
Existing pseudo-labeling methods have generated pseudo-labels when their confidence scores are higher than or equal to a certain threshold \cite{scarf, pseudo_label6} or a certain proportion \cite{proportion-self-training, curriculum_labeling}.
Although the cluster assumption is widely accepted for semi-supervised learning, naive confidence score based pseudo-labeling methods rarely guarantees the cluster assumption.
This assumption means that two close data points are likely to belong to the same cluster and the points in the same cluster are likely to the same class \cite{low_density, modified_cluster_assumption, pseudo_label}.
However, the current methods have little ability to distinguish whether or not pseudo-labels lie in high-density regions, i.e. clusters.

% These pseudo-labeling methods rarely guarantee the cluster assumption which is widely accepted for semi-supervised learning.
% The cluster assumption means that two close data points are likely to belong to the same cluster and the points in the same cluster are likely to the same class \cite{low_density, modified_cluster_assumption, pseudo_label}. 
% Naive confidence score based pseudo-labeling methods, however, have little ability to distinguish whether or not pseudo-labels lie in high-density regions, i.e. clusters.

To resolve this issue, we propose a new pseudo-labeling algorithm that is specialized for tabular data and generates reliable pseudo-labels which lie in high-density regions, using regularized confidence scores based on the likelihood of each pseudo-label, and demonstrate the superiority of the proposed algorithm with exhaustive experiments using various models and datasets.

We summarized our contributions as follows:
\begin{itemize}
\item We suggest the contemporary state-of-the-art pseudo-labeling method, curriculum pseudo-labeling, which has been used in the computer vision field so far, for tabular data.
\item We propose regularized pseudo-labeling which prefers pseudo-labels that lie in high-density regions to pseudo-labels in low-density regions.
\item We demonstrate that our proposed methods are compatible with various models and datasets. Our proposed approaches show better performance than existing methods in vigorous validation using extensive experiments.
\end{itemize}

\newpage

\begin{figure*}[ht]
    \vskip 0.2in
    \centering
    \begin{subfigure}[b]{0.3\textwidth}
        \centering
        \includegraphics[width=0.8\linewidth]{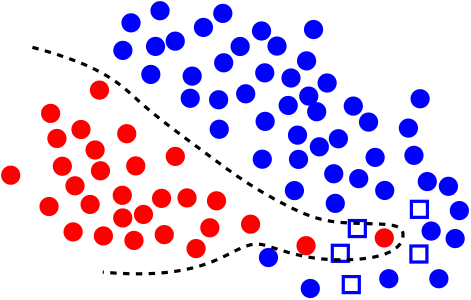}
        \caption{}
        \label{fig:illustration-a}
    \end{subfigure} \hfill
    \begin{subfigure}[b]{0.3\textwidth}
    \centering
        \includegraphics[width=0.8\linewidth]{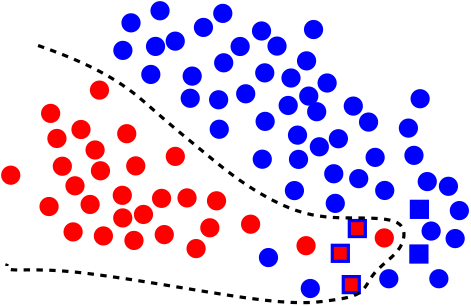}
        \caption{}
        \label{fig:illustration-b}
        \end{subfigure} \hfill
    \begin{subfigure}[b]{0.3\textwidth}
        \centering
        \includegraphics[width=0.8\linewidth]{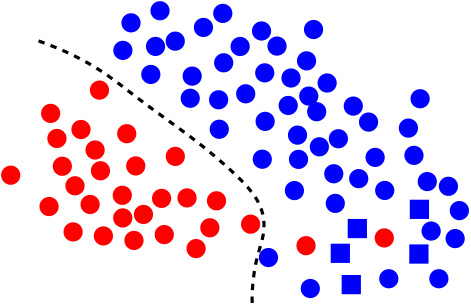}
        \caption{}
        \label{fig:illustration-c}
    \end{subfigure}
    \begin{adjustwidth}{-2cm}{-2cm}
    \caption{\textbf{Illustration for the existing issue of naive confidence score based self-training via visualization of the feature space.} The dashed line represents the decision boundary of the trained model, and red and blue for each class respectively. Unfilled squares mean unlabeled data of which ground truth class is blue. (a) The feature space of the model that is trained by supervised learning only, resulting in a biased region. (b) The feature space of the model that is trained by self-training with naive confidence scores, enhancing the biased region to become more biased than (a). (c) The feature space of the model that is trained by self-training with regularized confidence scores, generating more accurate pseudo-labels using likelihood, which leads to a more reliable decision boundary than (b). (See, the empirical result in Appendix \ref{appendix:analyes_pl})}
    \label{fig:illustration}
    \end{adjustwidth}
    \vskip -0.2in
\end{figure*}

\section{Related Work} \label{sec:related_work}

\subsection{Semi- and Self-Supervised Learning for tabular data} \label{related_work1}
To overcome those tabular datasets which scarce labels, several semi- and self-supervised learning methods have been proposed to minimize special loss such as reconstruction loss or contrastive loss \cite{vime, saint, subtab, tabnet}. 
There are also some models that can perform some pretraining procedures like masked language modeling (MLM) or replaced token detection (RTD) which are used for the language models \cite{bert, electra, tabtransformer, saint}. 
Although they have been successful to improve the performance of a given model, these methods have some constraints which require a special form of architecture such as an auto-encoder or transformer to use their special losses, or appropriate composition of datasets (e.g.\ categorical features to apply the pretraining procedures).
Despite these requirements, deep neural networks including transformers and auto-encoders have not been the dominant architecture when learning tabular data \cite{kaggle_2021_state, deep_learning_is_not_all_you_need} and the tabular data sometimes do not have any categorical features where pretraining procedures can be applied.

Some studies have attempted self-training as a semi-supervised learning method, but they do not pay much attention to self-training, just using it as a baseline for other methods.
Moreover, they have used out-of-date pseudo-labeling methods which adopt pseudo-labels that have a confidence score above a certain threshold or even adopt all pseudo-labels regardless of their values \cite{tabtransformer, scarf}.
Since it is hard to find an optimal threshold during the self-training cycle for different models and datasets, various studies in the computer vision field empirically show the inferiority of using a fixed threshold when generating pseudo-labels \cite{curriculum_labeling, flexmatch}.

\subsection{Curriculum Pseudo-Labeling} \label{related_work2}
Selecting the pseudo-labels to include in the training set is the most important component in self-training, and regarding pseudo-labeling, a number of methods have been proposed in computer vision field \cite{fixed, proportion-self-training, pseudo_label5, curriculum_labeling}. 
Among them, the state-of-the-art self-training method is curriculum pseudo-labeling \cite{curriculum_labeling}.
Recently, self-training seems to have been abandoned in favor of other methods like consistency regularization, but curriculum pseudo-labeling shows competitive results for various image datasets compared to the state-of-the-art non-self-training methods \cite{curriculum_labeling}.
Curriculum pseudo-labeling does not use a fixed threshold which is widely used in the tabular domain, but instead uses percentile scores to decide which samples to add.
Starting with pseudo-labels that have a top $r\%$ confidence score, curriculum pseudo-labeling increases the percentile by a certain amount during the self-training cycle until no unlabeled data is left.
In this paper, we tackle the convention of using only fixed threshold pseudo-labeling and empirically show the strength of curriculum pseudo-labeling for various models and datasets.

\subsection{Cluster Assumption} \label{related_work3}
A fundamental underlying assumption when training the classifier using a semi-supervised learning technique is the cluster assumption which states that the data samples in the same cluster have the same label and the decision boundary should not cross high-density regions, but instead, lie in low-density regions \cite{low_density, modified_cluster_assumption, pseudo_label}.
Hence, if the classifier follows the cluster assumption, the pseudo-label which lies in a high-density region is more reliable than those that lie in a low-density region.
However, the current pseudo-labeling methods which only use confidence scores do not guarantee that the pseudo-labels which have high confidence scores lie in high density regions.

\section{Methods}\label{sec:method}
In this section, we describe our novel pseudo-labeling strategy for tabular data.
We first define the self-training cycle that selects the subset of the pseudo-labels using a pseudo-labeler and iterates this process under a certain condition. %until neither performance gain is achieved nor unlabeled data is left.
Then, we demonstrate a regularized pseudo-labeling algorithm that replaces naive confidence scores with regularized confidence scores computed by our new scoring function $f$, which values pseudo-labels, in Algorithm \ref{alg:algorithm1} of Appendix \ref{appendix:algorithm}.

\subsection{Self-Training Cycle}\label{sec:self-training_cycle}

\textbf{Notation} Let $D_{L} = \{(x^{(i)}, y^{(i)})\}_{i = 1}^{N_{L}}$ be a labeled dataset of $N_{L}$ labeled samples 
where $x^{(i)}$ is the features of the $i^{th}$ sample and $y^{(i)}$ is the label of the $i^{th}$ sample.
Let $D_{U} = \{x^{(i)}, \varnothing\}_{i = 1}^{N_{U}}$ be an unlabeled dataset of $N_{U}$ samples for each of which has features $x^{(i)}$ only with no label. 
We also denote a subset of $D_{U}$ by $\Tilde{D}_{U}$, a subset of $N_{U}$ by $\Tilde{N}_{U}$, a set of features of $D_{U}$ by $X_{U}$, the scoring function by $f$, confidence score of input $x^{(i)}$ by $c^{(i)}$, and classifier by $C$.
For each unlabeled sample, pseudo-label $\Tilde{y}^{(i)}$ is generated by pseudo-labeler $\Phi_{l}(f, C, x^{(i)})$ (e.g.\ fixed threshold pseudo-labeling or curriculum pseudo-labeling) when the score of the pseudo label for given input $x^{(i)}$ is above a certain threshold, or satisfies a certain condition (e.g. the confidence score of input $x^{(i)}$ is in top $r$\%).
Let $\Tilde{D} = \{x^{(i)}, y^{(i)}$ or $\Tilde{y}^{(i)}\}_{i = 1}^{N_{L} + \Tilde{N}_{U}}$ a new training set which consists of $D_{L}$ and pseudo-labeled dataset $\Tilde{D}_{U}$ of $\Tilde{N}_{U}$ samples (note that unlabeled samples are not pseudo-labeled, if their scores for the pseudo-labeler do not meet the criteria).
Algorithm \ref{alg:algorithm1} shows the full pipeline of our self-training cycle.

\subsection{Regularized Pseudo-Labeling}\label{sec:regularized-pseudo-labeling}
The cluster assumption has been widely acknowledged when training a classifier by a semi-supervised learning technique \cite{low_density, modified_cluster_assumption, pseudo_label}.
Since the value of each pseudo-label is determined by confidence scores in previous pseudo-labeling approaches that do not consider whether a sample is located in a low-density region or high-density region, the cluster assumption has been often violated.
As shown in Figure \ref{fig:illustration}, if the classifier has a strong bias, there is a potential risk that the classifier generates a wrong high confidence score for a sample that lies in a low-density region but near the biased samples.
Our study is motivated by this limitation of existing pseudo-labeling approaches.

We assume that, as described in the cluster assumption, a classifier maps data into high-density regions according to each label, and its decision boundaries lie in low-density regions between the clusters.
Therefore, it is reasonable that if a pseudo-label lies in a high-density region, it would be more reliable. 
It is also natural that pseudo-labels that lie in high-density regions have higher likelihoods than those in low-density regions.

Based on the above assumption, we propose a regularized pseudo-labeling approach. 
To ensure the cluster assumption, we generate pseudo-labels for data samples that have both a high confidence score and a high likelihood by regularizing the confidence scores using their likelihoods.
To compute the likelihood of each pseudo-label, we assume that the location of the origin feature space is associated with the latent vector space where the decision boundaries lie. 
If the origin features of arbitrary data have a high likelihood for the label, the data also has a high likelihood for the label in the latent vector space, i.e.\ it lies in a high-density region in the latent vector space, not in a low-density region.
Note that, unlike image or text, each feature of tabular data has its own position in the table, so we can directly obtain the likelihood of each feature for all samples.

Let $x^{(i)} = \{x^{(i)}_{0}, x^{(i)}_{1}, ... , x^{(i)}_{m - 2}, x^{(i)}_{m - 1}\}$ which is composed of $m$ features, and
assume that each feature of $x^{(i)}$ is independent of each other.
The likelihood of the given $x^{(i)}$ for its pseudo-label $\Tilde{y} ^ {(i)}$  is determined as follows.

\begin{equation} \label{eq1}
    \begin{aligned}
        P(x^{(i)}|\Tilde{y}^{(i)}) 
         & = P(x^{(i)}_{0}, x^{(i)}_{1}, ... , x^{(i)}_{m - 2}, x^{(i)}_{m - 1}|\Tilde{y}^{(i)}) \\
            & = \frac{P(\Tilde{y}^{(i)} , x^{(i)}_{0},x^{(i)}_{1}, ... , x^{(i)}_{m - 2}, x^{(i)}_{m - 1})}{P(\Tilde{y}^{(i)})}
    \end{aligned}
\end{equation}

Then we apply the chain rule as follows.

\begin{equation} \label{eq2}
    \begin{aligned}
        & P(x^{(i)}|\Tilde{y}^{(i)}) \\
        & = \frac{P(\Tilde{y}^{(i)})P(x^{(i)}_{0}|\Tilde{y}^{(i)}), ..., P(x^{(i)}_{m - 1}|\Tilde{y}^{(i)},x^{(i)}_0,...,x^{(i)}_{m - 2})}{P(\Tilde{y}^{(i)})}   
    \end{aligned}
\end{equation}

Since we assume that the $m$ features of $x^{(i)}$ are independent of each other, 
we can simplify eq (\ref{eq2}) to eq (\ref{eq3}).

\begin{equation} \label{eq3}
    \begin{aligned}
        P(x^{(i)}|\Tilde{y}^{(i)}) 
        & = \frac{P(\Tilde{y}^{(i)})P(x^{(i)}_{0}|\Tilde{y}^{(i)}), ..., P(x^{(i)}_{m - 1}|\Tilde{y}^{(i)})}
                {P(\Tilde{y}^{(i)})} \\
        & = P(x^{(i)}_{0}|\Tilde{y}^{(i)})P(x^{(i)}_{1}|\Tilde{y}^{(i)}), ..., P(x^{(i)}_{m - 1}|\Tilde{y}^{(i)})
    \end{aligned}
\end{equation}

\begin{table*}[t]
\begin{adjustwidth}{-2cm}{-2cm}
\caption{Performance comparison of one supervised learning without self-training method (NONE) and four pseudo-labeling methods: FPL, R-FPL, CPL, and R-CPL using three binary classification datasets: 6 months mortality, 12 months reverse remodeling, and Albert with various models: XGBoost (XGB), LightGBM (LGBM), FT-Transformer (FTT), TabTransformer (TT), TabNet (TN), Saint (ST), and MLP.
(a) F1-scores of the methods for the 6 months mortality dataset. (b) F1-scores of the methods for the 12 months reverse remodeling dataset. (c) Accuracy scores of the methods for click-through rate prediction for a given advertisement in Albert dataset. (d) The average of the ranks for each method.
}
\label{table:1}
\end{adjustwidth}
\begin{adjustwidth}{-2cm}{-2cm}
\tiny
\vskip 0.15in
\begin{subtable}{0.5\linewidth}
\begin{center}
\begin{sc}
\scriptsize
\caption{}
\label{table:6m_mortality}
\begin{tabular}{l p{2em} p{2em} p{2em} p{2em} p{2em} p{2em} p{3em} r}
    \toprule
    Method & {\centering XGB} & {\centering LGBM} & {\centering FTT} & {\centering TT} & {\centering TN} &  {\centering ST} & {\centering MLP} \\
    \midrule
    None & 0.5534 & 0.5325 & 0.4544 & 0.4277 & 0.4482 & 0.5484 & 0.4249 \\ \hline
    FPL & 0.5565 & \textbf{0.5430}* & 0.4851 & \textbf{0.4468}* & 0.4518 & 0.5515 & 0.4478 \\
    R-FPL & 0.5630* & 0.5367 & 0.4863* & 0.4409 & \textbf{0.4652}* & \textbf{0.5779}* & 0.4626* \\ \hline
    CPL & 0.5479 & 0.5391* & 0.4833 & 0.4416 & 0.4593* & 0.5760 & 0.4606 \\
    R-CPL & \textbf{0.5681}* & 0.5379 & \textbf{0.4951}* & \textbf{0.4468}* & 0.4563 & 0.5776* & \textbf{0.4628}* \\
    \bottomrule
\end{tabular}
\end{sc}
\end{center}
\end{subtable} %
\begin{subtable} {0.5\textwidth}
\begin{center}
\begin{sc}
\scriptsize
\caption{}
\label{table:reverse_remodeling}
\begin{tabular}{l p{2em} p{2em} p{2em} p{2em} p{2em} p{2em} p{3em} r}
    \toprule
    Method & {\centering XGB} & {\centering LGBM} & {\centering FTT} & {\centering TT} & {\centering TN} &  {\centering ST} & {\centering MLP} \\
    \midrule
    None & 0.8184 & 0.8155 & 0.7447 & 0.7490 & 0.8168 & 0.8253 & 0.7467 \\ \hline
    FPL & 0.8166 & 0.8196 & 0.7453 & 0.7466 & 0.8170 & 0.8261 & 0.7525* \\
    R-FPL & 0.8187* & 0.8215* & \textbf{0.7570}* & 0.7492* & 0.8171* & 0.8281* & 0.7514 \\ \hline
    CPL & 0.8212 & 0.8207 & 0.7559* & \textbf{0.7507}* & \textbf{0.8175}* & \textbf{0.8285}* & 0.7526 \\
    R-CPL & \textbf{0.8219}* & \textbf{0.8218}* & 0.7555 & 0.7491 & 0.8172 & 0.8281 & \textbf{0.7555}* \\
    \bottomrule
\end{tabular}
\end{sc}
\end{center}
\end{subtable} %
\vskip -0.1in
\end{adjustwidth}

\begin{adjustwidth}{-1.5cm}{-1cm}
\vskip 0.15in
\begin{subtable}[t] {0.5\textwidth}
\begin{center}
\begin{sc}
\scriptsize
\caption{}
\label{table:albert}
\begin{tabular}{l p{2em} p{2em} p{2em} p{2em} p{2em} p{2em} p{3em} r}
    \toprule
    Method & {\centering XGB} & {\centering LGBM} & {\centering FTT} & {\centering TT} & {\centering TN} &  {\centering ST} & {\centering MLP} \\
    \midrule
    None & 0.6310 & 0.6570 & 0.5743 & 0.6026 & 0.6476 & 0.6542 & 0.5910 \\ \hline
    FPL & 0.6411 & 0.6476 & 0.5974 & 0.6134 & 0.6522* & 0.6579 & 0.6189 \\
    R-FPL & 0.6466* & \textbf{0.6560}* & 0.6276* & 0.6175* & 0.6499 & \textbf{0.6581}* & 0.6234* \\ \hline
    CPL & \textbf{0.6629}* & 0.6503 & \textbf{0.6366}* & 0.6186 & 0.6516 & 0.6568 & 0.6069 \\
    R-CPL & 0.6607 & 0.6516* & 0.6321 & \textbf{0.6190}* & \textbf{0.6526}* & 0.6575* & \textbf{0.6255}* \\
    \bottomrule
\end{tabular}
\end{sc}
\end{center}
\end{subtable} %
\hspace{1.5cm}
\begin{subtable}[t] {0.5\textwidth}
\begin{center}
\begin{sc}
\scriptsize
\caption{}
\label{table:model_total}
\begin{tabular}{l p{1em} p{2em} p{1em} p{1em} p{1em} p{1em} p{2em} | p{1.5em} r}
    \toprule
    Method & {\centering XGB} & {\centering LGBM} & {\centering FTT} & {\centering TT} & {\centering TN} &  {\centering ST} & {\centering MLP} & {\centering Avg} \\
    \midrule
    None & 4.3 & 3.7 & 5.0 & 4.7 & 5.0 & 5.0 & 5.0 & 4.7 \\ \hline
    FPL & 4.0 & 3.3 & 4.0 & 3.3 & 3.3 & 4.0 & 3.3 & 3.6 \\
    R-FPL & 2.7* & 2.7* & 2.0* & 3.0* & 2.7* & 2.0* & 2.7* & 2.5* \\ \hline
    CPL & 2.7 & 3.0 & 2.0 & 2.0 & 2.0 & 2.0 & 3.0 & 2.3 \\
    R-CPL & \textbf{1.3}* & \textbf{2.3}* & 2.0 & \textbf{1.7}* & 2.0 & 2.0 & \textbf{1.0}* & \textbf{1.7}* \\
    \bottomrule
\end{tabular}
\end{sc}
\end{center}
\end{subtable}
\end{adjustwidth}

\vskip -0.1in
\end{table*}

Lastly, for computational efficiency, we use a log-likelihood by applying logarithms on eq (\ref{eq3}). Because the log-likelihood is a negative value and we need a relative likelihood between pseudo-labels rather than an exact value, we scale the log-likelihood value using a min-max scaler to apply it easily to a scoring function $f$.
Then we formulate $f$ that uses the log-likelihood of the data for regularized pseudo-labeling as follows.
% Our new scoring function $f$ is as follows.

\begin{equation} \label{eq4}
    f(x) = (\alpha\gamma + 1)c/ (\alpha + 1)
\end{equation}

Where $\gamma$ is a log-likelihood for a sample,
$\alpha$ a hyperparameter for the log-likelihood that indicates how much the log-likelihood will affect the value of the pseudo-label, (where $\alpha \geq 0$),
and $c$ the confidence score of a trained classifier for the pseudo-label of unlabeled data $x$.

The higher value $\alpha$ becomes, the more influence regularized confidence score has on the pseudo-labeler. If $\alpha$ is 0, the pseudo-labeler only uses the confidence score to decide whether to generate the pseudo-label for given $x$. It is the same pseudo-labeler as the one used in previous studies. 
The reason we divide by $(\alpha + 1)$ in scoring function $f$ of eq (\ref{eq4}) is to make $f$ always return the value between 0 and 1 so that our new scoring function becomes compatible with the previous fixed threshold pseudo-labeling.

Now that the likelihood of each unlabeled sample can be cached before starting the training, only a few negligible operations are added during the self-training cycle.
Furthermore, since we use log-likelihood instead of likelihood, we can simply cache log-likelihood via not multiplication but summation.
Lastly, our modification on the self-training algorithm is only on how to generate pseudo-labels, thus it can be simply applied to any self-training algorithms with low computation (See Appendix \ref{appendix:time_overhead}). 

\section{Results and Discussion}\label{sec:experiments}

\begin{table*}[t]
\begin{adjustwidth}{-2cm}{-2cm}
\caption{Performance comparison of one supervised learning without self-training method (NONE) and four pseudo-labeling methods: FPL, R-FPL, CPL, and R-CPL using seven binary classification datasets and five multiclass classification datasets with XGBoost.
(a) Performance comparison of the methods for seven binary classification datasets: Christine, Jasmine, Madeline, Philippine, and Sylvine in OpenML, and Coupon, and Bank Marketing (Bank) in UCI Machine Learning Repository. The performance metric used for the first six datasets is accuracy score and for Bank dataset F1-score. (b) Performance comparison of the methods for five multiclass classification datasets: Dilbert, Fabert, Splice, MNIST, and Steel Plates Fault (Steel) in OpenML. The performance metric used for the first four datasets is balanced accuracy score and for Steel dataset F1-score. (c) The average of the ranks for each method.
}
\label{table:2}
\tiny
\vskip 0.15in
\begin{center}
\begin{subtable}{1.0\textwidth}
\begin{center}
\begin{sc}
\small
\caption{}
\label{table:binary}
\begin{tabular}{lcccccccr}
    \toprule
    Method & {\centering Christine} & {\centering Jasmine} & {\centering Madeline} & {\centering Philippine} & {Sylvine} &  {\centering Coupon} & {\centering Bank} \\
    \midrule
    None & 0.7071 & 0.7922 & 0.6930 & 0.7219 & 0.9290 & 0.7142 & 0.5635 \\ \hline
    FPL & \textbf{0.7148}* & 0.7932 & 0.7080 & 0.7305* & 0.9295* & 0.7143 & 0.5676 \\
    R-FPL & 0.7086 & 0.7989* & 0.7236* & 0.7274 & 0.9290 & 0.7215* & 0.5711* \\ \hline
    CPL & 0.7111* & 0.7969 & 0.7080 & 0.7250 & 0.9297 & 0.7095 & 0.5707 \\
    R-CPL & 0.7082 & \textbf{0.7993}* & \textbf{0.7261}* & \textbf{0.7311}* & \textbf{0.9301}* & \textbf{0.7250}* & \textbf{0.5730}* \\
    \bottomrule
\end{tabular}
\end{sc}
\end{center}
\end{subtable}
\end{center}
\vskip 0.15in
\hspace{1cm}
\begin{subtable} {0.55\textwidth}
\begin{center}
\begin{sc}
\small
\caption{}
\label{table:multi}
\noindent %\hspace*{2em}%
\begin{tabular}{lcccccr}
    \toprule
    Method & {\centering Dilbert} & {\centering Fabert} & {\centering Splice} & {\centering MNIST} & {\centering Steel} \\
    \midrule
    None & 0.9581 & 0.5872 & 0.9464 & 0.9639 & 0.7228  \\ \hline
    FPL & 0.9639 & 0.5870 & 0.9529 & 0.9645 & 0.7269 \\
    R-FPL & \textbf{0.9646}* & \textbf{0.5890}* & \textbf{0.9545}* & \textbf{0.9647}* & 0.7300* \\ \hline
    CPL & 0.9633 & 0.5875 & 0.9503 & 0.9644 & 0.7316 \\
    R-CPL & 0.9642* & 0.5881* & 0.9531* & 0.9646* & \textbf{0.7351}* \\
    \bottomrule
\end{tabular}
\end{sc}
\end{center}
\end{subtable}%
\noindent \hspace*{10em}%
\begin{subtable}{0.35\textwidth}
\begin{center}
\begin{sc}
\small
\caption{}
\label{table:bin_mul_rank}
\begin{tabular}{lcc|cr}
    \toprule
    Method & {\centering Binary} & {\centering Multi-class} & {\centering Avg} \\
    \midrule
    None & 4.7 & 4.8 & 4.8 \\ \hline
    FPL & 2.9 & 3.6 & 3.3 \\
    R-FPL & 2.6* & \textbf{1.4}* & 2.0* \\ \hline
    CPL & 3.1 & 3.4 & 3.3 \\
    R-CPL & \textbf{1.4}* & 1.8* & \textbf{1.6}* \\
    \bottomrule
\end{tabular}
\end{sc}
\end{center}
\end{subtable}
\vskip -0.1in
\end{adjustwidth}
\end{table*}

We compared the performance of curriculum pseudo-labeling and regularized pseudo-labeling with existing methods from two aspects; unconstrained to models and unconstrained to datasets via experiments using various models on diverse tabular datasets.
Further, we conducted additional experiments to compare self-training and pretraining which is a widely used self-supervised learning method to leverage unlabeled data for tabular data, and to observe the performance for labeled samples of various sizes.
In all these experiments for performance validation, we conducted supervised learning without self-training (NONE), fixed threshold pseudo-labeling (FPL), regularized fixed threshold pseudo-labeling (R-FPL), curriculum pseudo-labeling (CPL), and regularized curriculum pseudo-labeling (R-CPL) with limited training samples to demonstrate the superiority of our proposed methods: curriculum pseudo-labeling and regularized pseudo-labeling.

\subsection{Datasets}\label{sec:datasets}

We used ten binary classification datasets and five multiclass classification datasets in this study.
These include two binary classification datasets: 6 months mortality, and 12 months reverse remodeling from KAMIR (Korea Acute Myocardial Infarction Registry, which has follow-up data of patients after acute myocardial infarction),
six binary classification datasets: Albert, Christine, Jasmine, Madeline, Philippine, and Sylvine from OpenML which is an online platform that provides enormous amounts of datasets available in public, two binary classification datasets: Bank Marketing, and In-Vehicle Coupon Recommendation from UCI Machine Learning Repository which is another open dataset repository, and five multiclassification datasets: Dilbert, Fabert, Splice, MNIST, and Steel Plates Fault from OpenML. 

For the 6 months mortality, the 12 months reverse remodeling, Bank Marketing, and Steel Plates Fault datasets that are imbalanced (i.e. labels are skewed), we measured F1-score, and for the rest, accuracy score.

Note that, in contrast to the 6 months mortality dataset, the 12 months reverse remodeling dataset requires 6 months longer follow-ups, and extra examinations to measure the heart function.
Both consist of the same patients. 
While the 6 months mortality dataset has labels only for all 15,628 patients, the 12 months reverse remodeling dataset has labels for 5,742 patients, and the rest of the patients are unlabeled. 
This motivated us to study semi-supervised learning for tabular datasets.

The details of the datasets are described in Appendix \ref{appendix_dataset}.

\begin{figure*}[t]
    \vskip 0.11in
    \centering
    \begin{subfigure}[b]{1.0\textwidth}
        \centering
        \includegraphics[width=.32\linewidth]{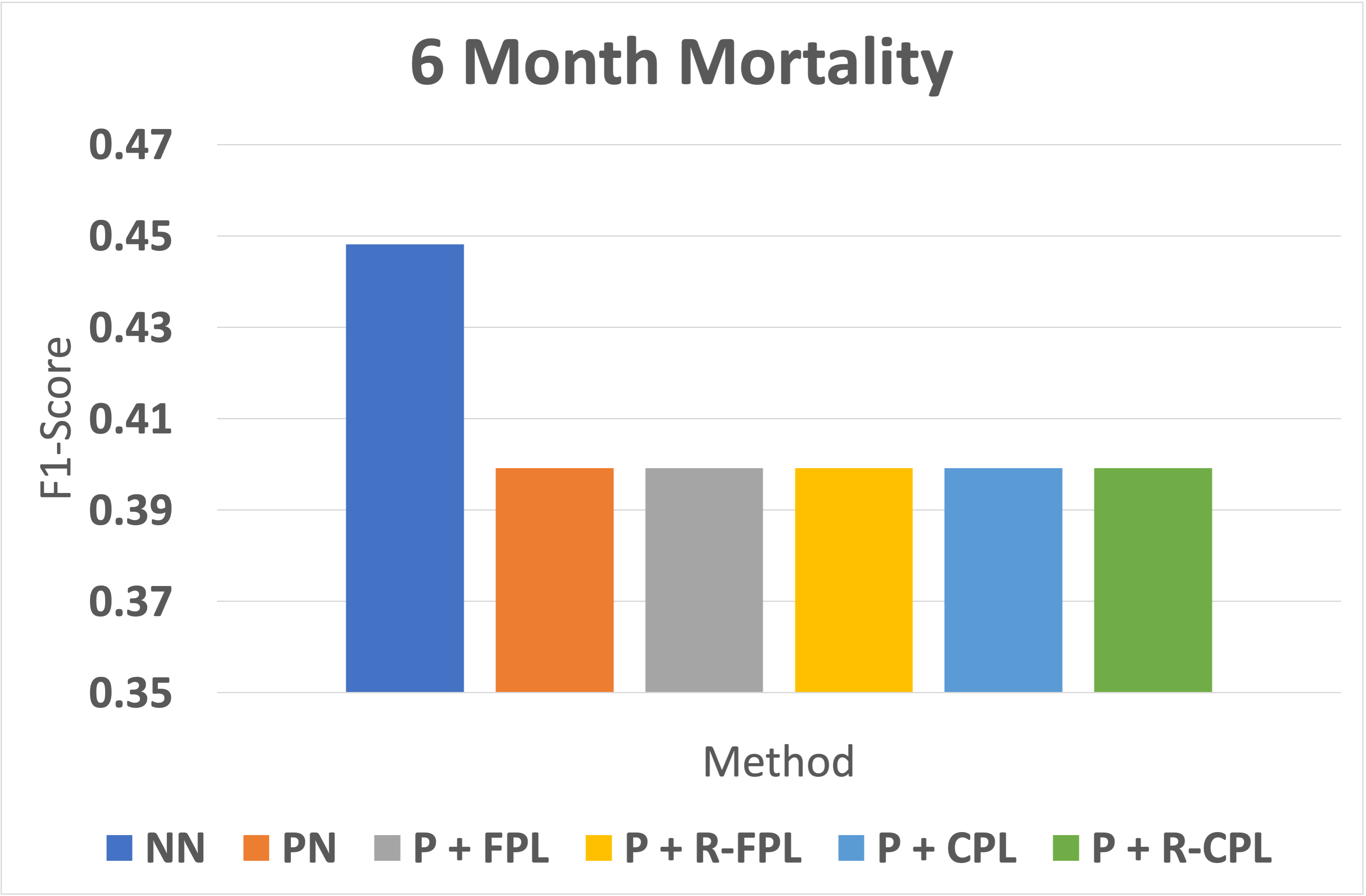}%
        \hfill
        \includegraphics[width=.32\linewidth]{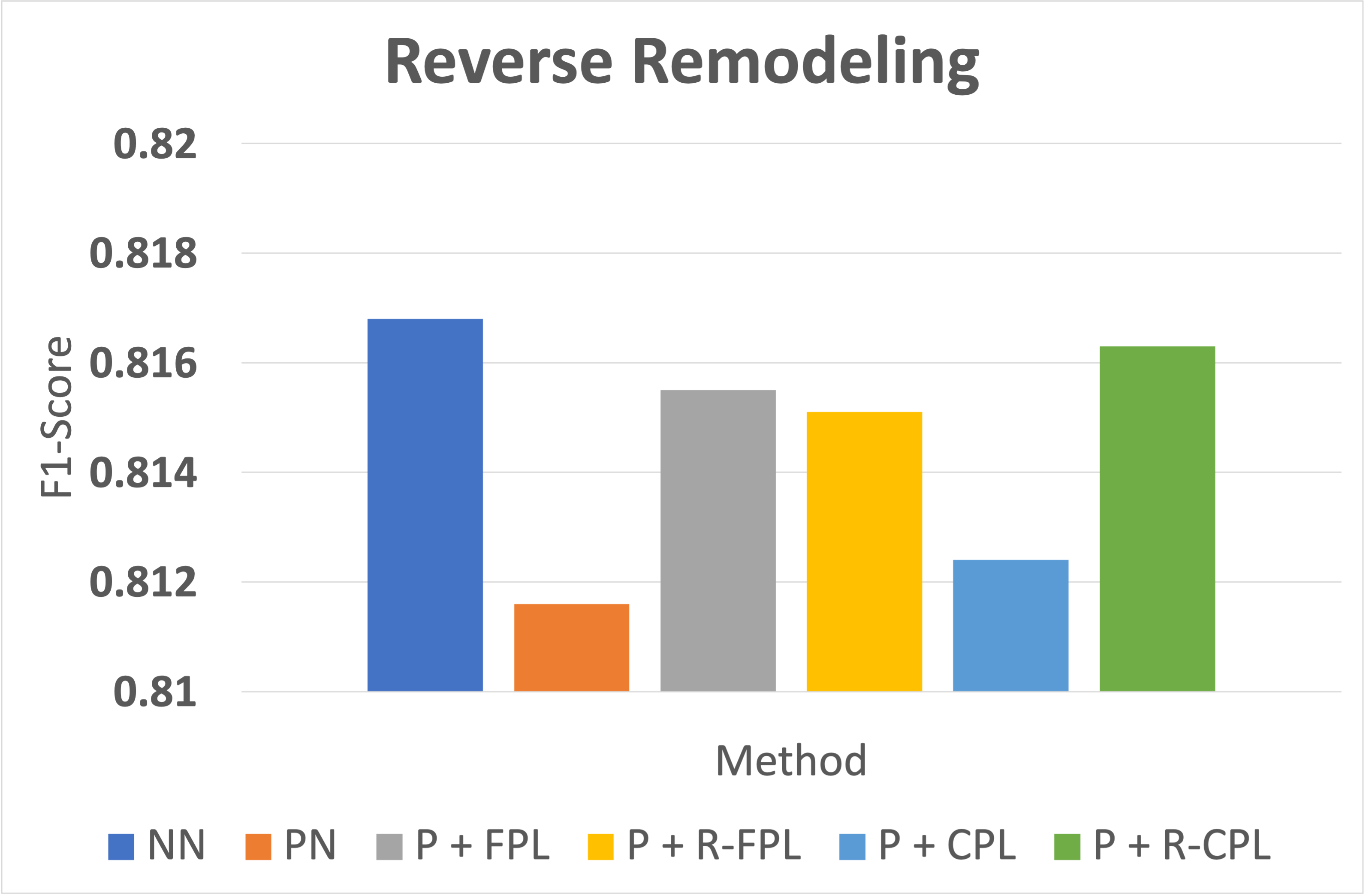}
        \hfill
        \includegraphics[width=.32\linewidth]{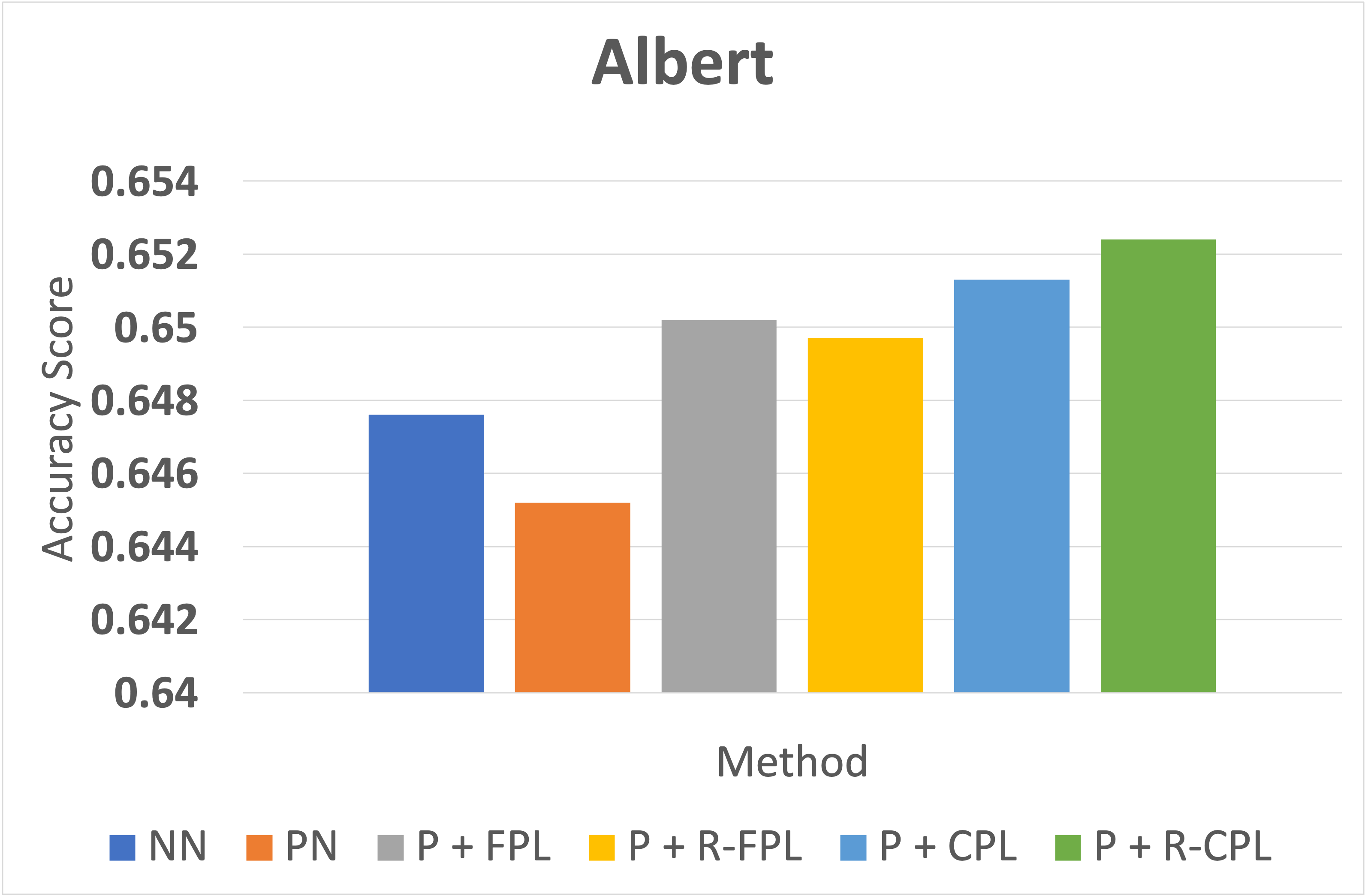}
        \caption{TabNet}
        \label{fig:fig2-a}
    \end{subfigure} %\hfill
    \begin{subfigure}[b]{1.0\textwidth}
        \centering
        \includegraphics[width=.32\linewidth]{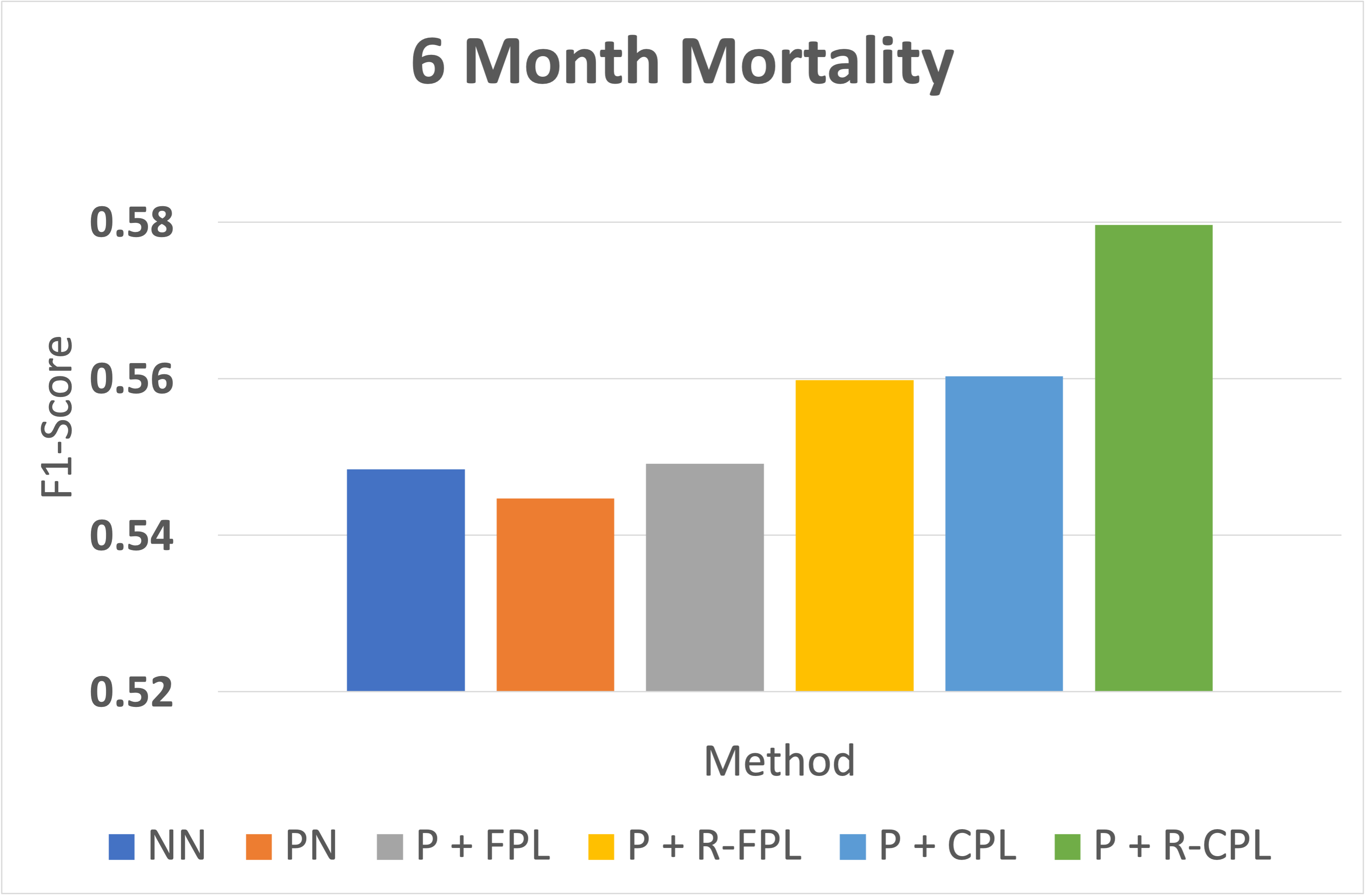}%
        \hfill
        \includegraphics[width=.32\linewidth]{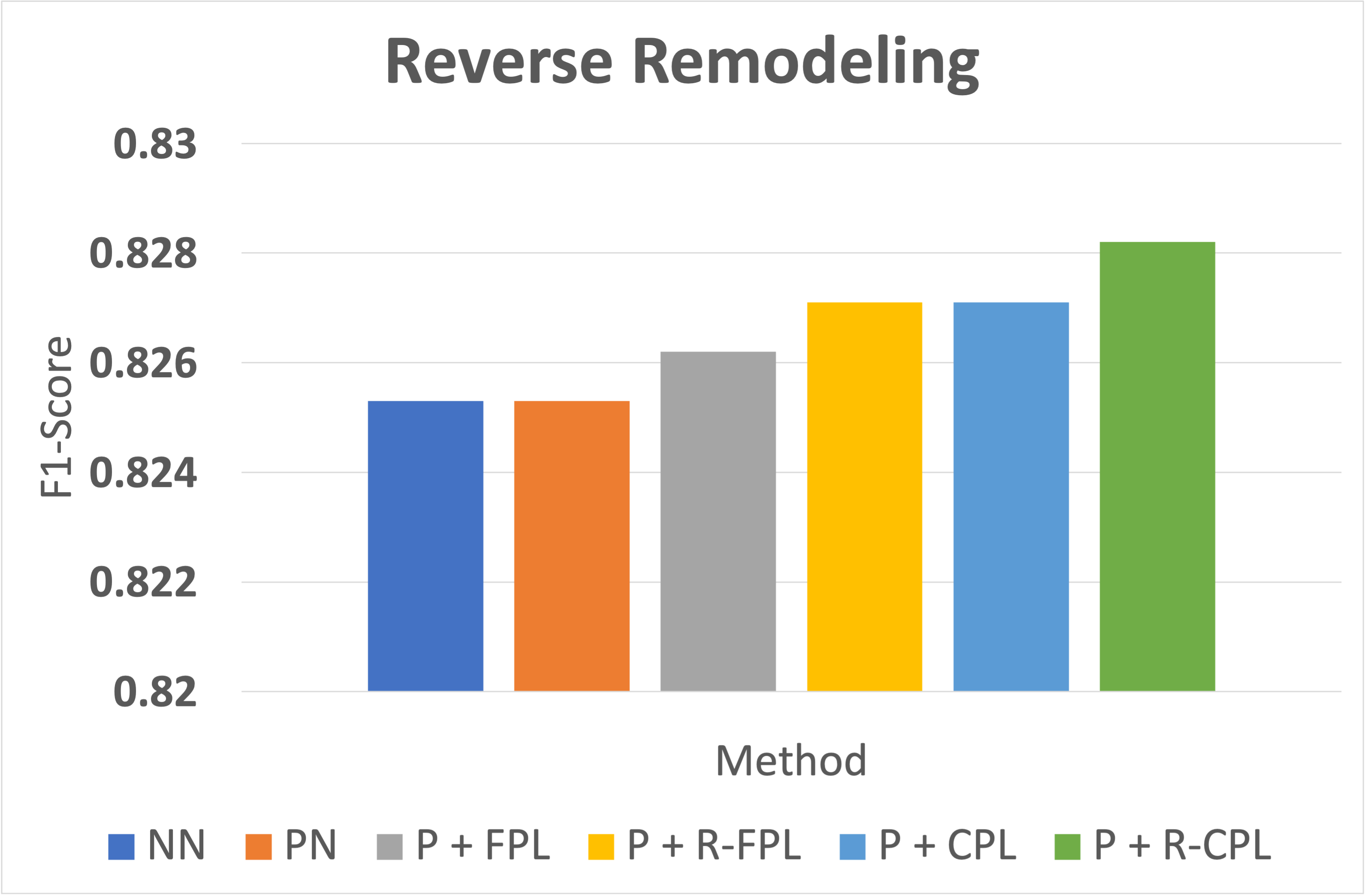}
        \hfill
        \includegraphics[width=.32\linewidth]{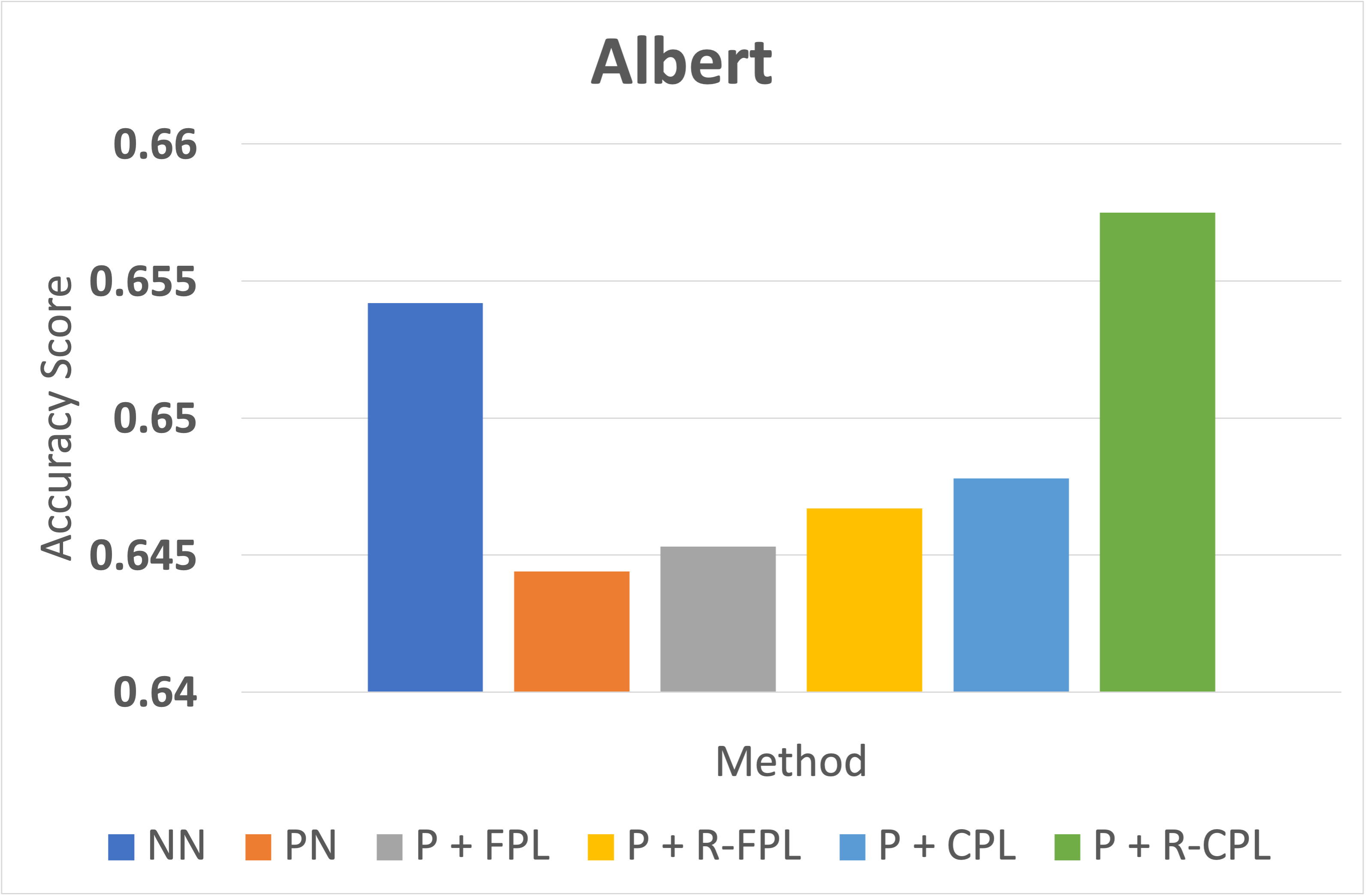}
        \caption{Saint}
        \label{fig:fig2-b}
    \end{subfigure} %\hfill
    \begin{adjustwidth}{-2cm}{-2cm}
    \caption{Performance comparison of one supervised learning method which used neither pretraining nor self-training (NN), one pretraining method without self-training (PN), and four pretraining + self-training methods: P + FPL, P + R-FPL, P + CPL, and P + R-CPL on three binary classification datasets: 6 months mortality, 12 months reverse remodeling, and Albert with TabNet and Saint. The performance metric for the 6 months mortality and the reverse remodeling is F1-score and for Albert is accuracy score.}
    \label{fig:fig2}
    \end{adjustwidth}
    \vskip -0.2in
\end{figure*}

\subsection{Implementation}\label{sec:implementation}
We implemented our approaches as follows.
We set the threshold of the fixed threshold pseudo-labeling as 0.6 empirically. For the curriculum pseudo-labeling, we set the initial threshold as 20\% and the stepping threshold percent as 20\% as performed in \cite{curriculum_labeling}.
During the self-training cycle, we reinitialized all classifiers to alleviate the accumulation of confirmation bias after generating pseudo-labels \cite{curriculum_labeling}.
For early stopping, we set 50 early stop patience rounds for GBDTs and 50 early stop patience epochs for neural networks.
For hyperparameter tuning, we used Optuna which is a hyperparameter optimization framework \cite{optuna} with 100 trials for all models except Saint of which hyperparameters are optimized with 30 trials since Saint requires a vast amount of training time.

For the likelihood of pseudo-labels, we selected a set of important features using Boruta-SHAP \cite{boruta_shap}. This can reduce time required to calculate the likelihood (note that the 
 computation of the likelihood for pseudo-labels needs a linear time complexity proportional to the dimension of features in data). For discrete features, we computed the likelihood of each discrete value based on the empirical distribution. For continuous features, we digitized to 10 discrete values and then used the empirical distribution.
For hyperparameter $\alpha$ in eq \ref{eq4}, which signifies the influence of the likelihood, we determined the value of $\alpha$ using grid search from 0.25 to 1.0 of which an interval is 0.25.
All experiments were performed using 3-fold stratified cross-validation with random seed 0.
To meet the constraint of KAMIR, since the reverse remodeling dataset of KAMIR includes only roughly 30\% of the samples labeled, we limited training to approximately 30\% of training samples for the other datasets (except Albert, Dilbert, and Bank Marketing data) and treated the rest as unlabeled. 
More details of our implementation are described in Appendix \ref{appendix_dataset} and \ref{appendix_hyperparameter}.

\subsection{Performance evaluation}\label{sec:results}

We validated the performance of our proposed approaches using several different models and tabular datasets.
To evaluate the performance of our proposed pseudo-labeling methods with different models, we conducted experiments for three binary classification datasets using two gradient boosting decision tree models (GBDT): XGBoost \cite{xgboost}, and LightGBM \cite{lightgbm}, four transformers: TabTrasformer \cite{tabtransformer}, FT-Transformer \cite{ft-transformer}, TabNet \cite{tabnet}, and Saint \cite{saint}, and one multilayer perceptron (MLP) with an embedding layer for categorical features, and report the results in Table \ref{table:1}.
To observe the performance of our proposed methods on different datasets, we built models for seven binary classification datasets and five multiclass classification datasets using XGBoost which is commonly used for tabular data, and show the results in Table \ref{table:2}.
For Saint and TabNet which include a pretraining step that can work with any tabular dataset, we also compared the performance of one pretraining-only method with one supervised-only method, and four self-training methods of the pretrained model, in Figure \ref{fig:fig2}.
Further, we conducted additional experiments with XGBoost and Saint to observe the performance for labeled samples of various sizes, and present the results in Table \ref{table:3} and Table \ref{table:1a}.
% Table \ref{table:1} shows the results of a set of different models, and Table \ref{table:2}  the results of a set of different datasets with XGBoost.
%Figure \ref{fig:fig2} illustrates the effect of self-training with pretraining which is a widely used self-supervised learning method for tabular data using TabNet and Saint that enable pretraining.
% We also validated the performance of our proposed methods in different labeled sample sizes using XGBoost and Saint (see Table \ref{table:3} and Table \ref{table:1a}). 
We marked an asterisk for the better one between the existing pseudo-labeling and our regularized pseudo-labeling methods, and the best score in bold in each column in all the tables.
To summarize the performance comparison in Tables \ref{table:1} and \ref{table:2}, we ranked the methods in each column and computed the average rank of the methods (see Table \ref{table:model_total}, and Table \ref{table:bin_mul_rank}, and note that the lowest rank is the best).

\begin{table*}[t]
\begin{adjustwidth}{-2cm}{-2cm}
\caption{Performance comparison of one supervised learning without self-training method (NONE), and four pseudo-labeling methods: FPL, R-FPL, CPL, R-CPL using three datasets: 6 months mortality, 12 months reverse remodeling, and Albert with XGBoost and Saint, depending on the number of labeled data. As an evaluation metric for the 6 months mortality, and the reverse remodeling, we used F1-Score and, for Albert, accuracy score.
For Saint, we also conducted pretraining as well.
(a) Performance comparison of the methods with XGBoost (b) Performance comparison of the methods with Saint.
}
\label{table:3}
\vskip 0.11in
\hspace{0.5cm}
\begin{subtable}[t] {\textwidth}
\begin{center}
\begin{sc}
\tiny
\caption{}
\label{table:3_XGB}
\begin{tabular}{l|cccc|cccc|ccccr}
    \toprule
     & \multicolumn{4}{c|}{\centering 6M Mortality} & \multicolumn{4}{c|}{Reverse Remodeling} & \multicolumn{4}{c}{Albert} \\ 
    \hline
    Method $\backslash$ N Labels & 500 & 1000 & 2000 & 3000 & 500 & 1000 & 2000 & 3828 & 1000 & 3000 & 5000 & 7000 \\ 
    \midrule
    None & 0.3646 & 0.4503 & 0.5127 & 0.5534 & 0.8066 & 0.8143 & 0.8155 & 0.8184 & 0.6050 & 0.6143 & 0.6234 & 0.6310 \\ \hline
    FPL & 0.4000* & 0.4625 & 0.5144 & 0.5565 & 0.8092 & 0.8183 & 0.8187* & 0.8166 & 0.6150 & 0.6272 & 0.6317* & 0.6411 \\
    R-FPL & 0.3951 & 0.4757* & \textbf{0.5287}* & 0.5630* & \textbf{0.8139}* & 0.8186* & 0.8172 & 0.8187* & \textbf{0.6193}* & 0.6276* & 0.6281 & 0.6466* \\ \hline
    CPL & 0.4047 & 0.4664 & 0.5207 & 0.5479 & 0.8087 & \textbf{0.8194}* & \textbf{0.8188}* & 0.8212 & 0.6115 & 0.6308 & \textbf{0.6361}* & \textbf{0.6629}* \\
    R-CPL & \textbf{0.4071}* & \textbf{0.4809}* & 0.5257* & \textbf{0.5681}* & 0.8121* & 0.8184 & 0.8186 & \textbf{0.8219}* & 0.6173* & \textbf{0.6321}* & 0.6339 & 0.6607 \\
    \bottomrule
\end{tabular}
\end{sc}
\end{center}
\end{subtable}
\vskip -0.1in
\vskip 0.15in
\hspace{0.5cm}
\begin{subtable}[t] {\textwidth} 
\begin{center}
\begin{sc}
\tiny
\caption{}
\label{table:3_Saint}
\begin{tabular}{l|cccc|cccc|ccccr}
    \toprule
     & \multicolumn{4}{c|}{\centering 6M Mortality} & \multicolumn{4}{c|}{Reverse Remodeling} & \multicolumn{4}{c}{Albert} \\ 
    \hline
    Method $\backslash$ N Labels & 500 & 1000 & 2000 & 3000 & 500 & 1000 & 2000 & 3828 & 1000 & 3000 & 5000 & 7000 \\ 
    \midrule
    None & 0.3644 & 0.4614 & 0.5287 & 0.5484 & 0.8148 & 0.8176 & 0.8207 & 0.8253 & 0.6067 & 0.6485 & 0.6519 & 0.6542 \\ \hline
    FPL & 0.4078 & 0.5024 & \textbf{0.5627}* & 0.5515 & 0.8193 & 0.8193 & 0.8244 & 0.8261 & 0.6359 & 0.6495 & 0.6519 & 0.6579 \\
    R-FPL & \textbf{0.4705}* & \textbf{0.5307}* & 0.5621 & \textbf{0.5779}* & 0.8197* & 0.8213* & 0.8247* & 0.8281* & \textbf{0.6400}* & \textbf{0.6527}* & \textbf{0.6549}* & \textbf{0.6581}* \\ \hline
    CPL & 0.4277 & 0.5136 & 0.5400 & 0.5760  & 0.8182 & 0.8195 & 0.8249 & \textbf{0.8285}* & 0.6207 & 0.6507 & 0.6522 & 0.6568 \\
    R-CPL & 0.4469* & 0.5284* & 0.5516* & 0.5776* & \textbf{0.8203}* & \textbf{0.8242}* & \textbf{0.8263}* & 0.8281 & 0.6278* & 0.6512* & 0.6531* & 0.6575* \\ \hline
    Pretraining & 0.2651 & 0.4850 & 0.2651 & 0.5447 & 0.8175 & 0.8192 & 0.8211 & 0.8258 & 0.6172 & 0.6385 & 0.6420 & 0.6467 \\
    \bottomrule
\end{tabular}
\end{sc}
\end{center}
\end{subtable} %
\vskip -0.1in
\end{adjustwidth}
\end{table*}

\subsubsection{Performance comparison in various models}
We evaluated the performance of our methods in various models using three datasets.
For all the models, \textbf{regularized pseudo-labeling generally outperforms non-regularized ones} (see Table \ref{table:1}).
Specifically, all of R-FPL's ranks exceed FPL's ranks, and all of R-CPL's ranks are higher or equal to CPL's (Table \ref{table:model_total}).
Interestingly CPL's ranks tend to perform better than FPL's by a large margin, but regularized pseudo-labeling makes FPL become competitive with CPL.
Moreover, CPL shows the lowest rank with regularized pseudo-labeling in most cases.
% We suggest choosing a better one between CPL and FPL with our regularized pseudo-labeling method.

\subsubsection{Performance comparison in various datasets}
To show the performance of our methods on various datasets, we experimented using various datasets: seven datasets for binary classification and five for multiclass classification (See Table \ref{table:2}).
We used XGBoost in this validation, which is one of the most commonly used models for tabular data.
In this experiment, \textbf{the average rank of regularized pseudo-labeling exceeds its non-regularized counterpart} (Table \ref{table:bin_mul_rank}).
To be specific, for the binary classification tasks, R-CPLs outperform other methods except Christine dataset, and R-FPLs surpass FPLs in the majority (Table \ref{table:binary}). 
In the multiclass classification, all the regularized pseudo-labeling methods outperform their non-regularized pseudo-labeling counterpart, and R-FPLs are usually the best (Table \ref{table:multi}).
Note that \textbf{R-CPL show consistently the lowest rank for the binary classification tasks} as illustrated in Table \ref{table:1} and Table \ref{table:2}.

\subsubsection{Self-training and pretraining}\label{sec:4.3.3}
In Figure \ref{fig:fig2}, we compared the effect of self-supervised pretraining in the tabular domain with self-training.
\textbf{While the pretrained TabNet and Saint fail to improve or even hurt the performance of supervised learning only, self-training always boosts the performance of the models even with pretraining} except for the 6 months mortality dataset with the pretrained TabNet, where the pretrained model might fail to generate useful pseudo-labels.
These results are in line with the computer vision field where pretraining shows worse results than self-training \cite{rethinking_pretraining}.
Further, the 6 months mortality dataset includes positive samples of only 3\%, which can be regarded as a highly skewed dataset. We suspect that this skewed data could cause the poor results of the pretraining for the 6 months mortality dataset.
To resolve this question, we further experimented in Section \ref{sec:4.3.4}.

\subsubsection{Performance comparison in labeled samples of various sizes}\label{sec:4.3.4}
Training labeled samples of various sizes is one of the common settings for evaluating semi- and self-supervised learning methods.
We observed the performance patterns of our self-training methods with labeled samples of various sizes using three datasets with XGBoost and Saint respectively in Table \ref{table:3}.
In most cases, \textbf{we obtained more performance gain with the lower number of labeled examples using the self-training methods, especially with our regularized pseudo-labeling} (See Figure \ref{fig:fig2a}).
For instance, on the 6 months mortality dataset using Saint with 500 labels, R-FPL and R-CPL achieve 29\% and 23\% performance gain respectively.

We conducted extra experiments in Table \ref{table:3_Saint} and Table \ref{table:1a} to confirm the performance of pretraining on the skewed datasets (Note the Bank Marketing dataset which is used in Table \ref{table:1a} is another skewed dataset that has only 12\% positive samples), and verified that pretraining has a potential problem with the skewed datasets.
In these experiments, pretraining sometimes greatly degrades performance when the datasets are skewed and more limited (see the results of the 6 months mortality data in Table \ref{table:3_Saint} and Table \ref{table:1a}).
In addition, these experiments, in the low-labeled data regimes, show that pretraining for Saint usually tend to achieve some performance gains, but self-training achieves much more performance improvement, which is also lined with the results of previous work in computer vision \cite{rethinking_pretraining}. 

\section{Limitations}

% \subsection{Likelihood computation}
Although our regularized pseudo-labeling contributes performance enhancement in the tabular domain, there is still room to improve. 
Since we use the empirical distribution, the characteristics of the individual features in given datasets may not be fully reflected in the calculation of the likelihood. 
This could explain why no performance gain in some experiments such as on the Christine dataset in Table \ref{table:binary}. 
If various feature characteristics are considered in the likelihood calculation, the accuracy of our regularized pseudo-labeling could be improved.

\section{Conclusion}\label{sec:conclusion}

In this paper, we revisit self-training which is a universal semi-supervised learning method for tabular data and guarantees performance enhancement.
Furthermore, we tackle the current conventions which use fixed threshold pseudo-labeling, and naive confidence score based pseudo-labelings that violate the cluster assumption.
Instead, we introduce curriculum pseudo-labeling for tabular data which is the state-of-the-art pseudo-labeling in the computer vision field, and show its strength.
In addition, we propose a regularized pseudo-labeling approach that uses a regularized confidence score that can guarantee the cluster assumption by generating pseudo-labels that have high confidence scores and likelihoods.
We highlight that our regularized pseudo-labeling outperforms for most tabular datasets and models with rigorous evaluation, and is easily applied with negligible overhead to any self-training algorithms.

\bibliography{preprint}

\newpage
\appendix
\onecolumn

\setcounter{table}{0}
\renewcommand{\thetable}{A\arabic{table}}

\setcounter{figure}{0}
\renewcommand{\thefigure}{A\arabic{figure}}

\setcounter{algorithm}{0}
\renewcommand{\thealgorithm}{A\arabic{algorithm}}

\newgeometry{textwidth=18cm, textheight=25cm}
\section{Supplementaries} \label{appendix:supple}
\subsection{Analyses of Pseudo-Labels} \label{appendix:analyes_pl}
% \vskip -3.5in
\begin{figure*}[ht]
    \vskip 0.2in
    \centering
    \begin{subfigure}[b]{0.45\textwidth}
        \centering
        \includegraphics[width=0.8\linewidth]{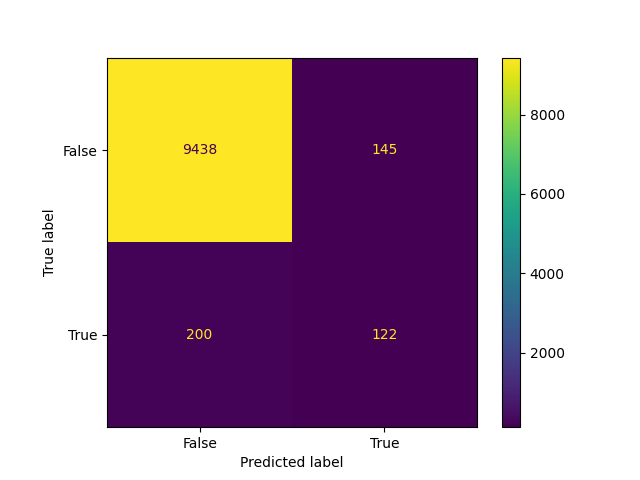}
        \caption{Non-Regularized}
        \label{fig:pl-a}
    \end{subfigure} \hfill
    \begin{subfigure}[b]{0.45\textwidth}
    \centering
        \includegraphics[width=0.8\linewidth]{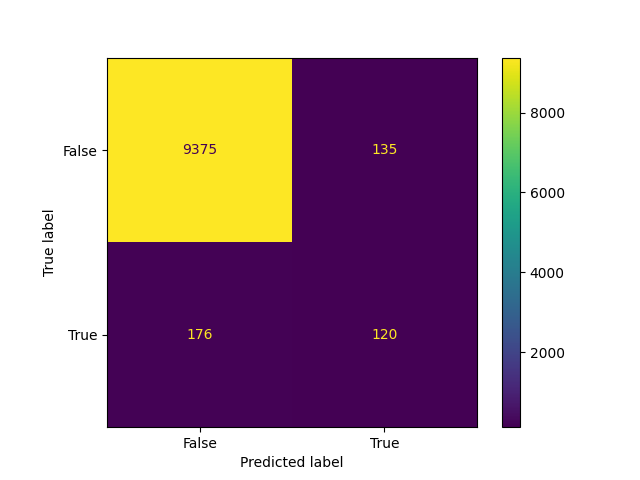}
        \caption{Regularized}
        \label{fig:pl-b}
    \end{subfigure} \hfill
    \label{fig:1a_pl}
\caption{Confusion matrix of pseudo-labels generated by fixed threshold pseudo-labeling on the 6 months mortality dataset using Saint with 500 labels. (a) Confusion matrix of non-regularized fixed threshold pseudo-labeling. (b) Confusion matrix of regularized fixed threshold pseudo-labeling.}
\vskip -0.2in
\end{figure*}
%\vskip -6in
\begin{table}[H]
\caption{Evaluation metrics of pseudo-labels generated by fixed threshold pseudo-labeling on the 6 months mortality dataset using Saint with 500 labels. Note that they used the same classifier, but show a different result.}
\label{table:pl}
\vskip 0.15in
\begin{center}
%\begin{scriptsize}
\begin{sc}
\begin{tabular}{lccccr}
\toprule
Method $\backslash$ Metric & Precision & Recall & F1-Score \\
\midrule
Non-Regularized & 0.4569 & 0.3789 & 0.4143 \\
Regularized & 0.4706 & 0.4054 & 0.4356 \\
\bottomrule
\end{tabular}
\end{sc}
%\end{scriptsize}
\end{center}
\vskip -0.1in
\end{table}
% \vskip 50in

\subsection{Time Overhead Comparison} \label{appendix:time_overhead}
\begin{table}[H]
 \caption{Time overhead (sec) comparison of pseudo-labeling step between FPL and R-FPL using XGBoost. 6 MM denotes the 6 months mortality dataset and 12 MRR denotes the 12 months reverse remodeling dataset.}
\label{table:time_overhead}
\vskip 0.15in
\begin{center}
\begin{tiny}
\begin{sc}
\begin{adjustwidth}{-0.75cm}{-1cm}
\begin{tabular}{lp{0.04\textwidth}p{0.04\textwidth}p{0.04\textwidth}p{0.04\textwidth}p{0.04\textwidth}p{0.04\textwidth}p{0.04\textwidth}p{0.04\textwidth} p{0.04\textwidth}p{0.04\textwidth}p{0.04\textwidth}p{0.04\textwidth}p{0.04\textwidth}p{0.04\textwidth}p{0.04\textwidth}r}
\toprule
Method $\backslash$ Dataset & 6MM & 12MRR & Albert & Christine & Jasmine & Madeline & Philippine & Sylvine & Coupon & Bank & Dilbert & Fabert & Splice & MNIST & Steel \\
\midrule
FPL & 0.05981 & 0.04338 & 0.03253 & 0.12931 & 0.10782 & 0.02810 & 0.03495 & 0.01537 & 0.02137 & 0.06598 & 0.27194 & 0.27571 & 0.08851 & 1.34567 & 0.04935 \\
R-FPL & 0.13031 & 0.14369 & 0.03419 & 0.15203 & 0.11831 & 0.03535 & 0.03740 & 0.02706 & 0.02386 & 0.14330 & 0.31172 & 0.33462 & 0.11316 & 1.42859 & 0.01252 \\
\bottomrule
\end{tabular}
\end{adjustwidth}
\end{sc}
\end{tiny}
\end{center}
\vskip -0.1in
\end{table}

\newpage

\subsection{Performance Comparison in Labeled Samples of Various Sizes}\label{appendix:label_sizes}

\begin{figure*}[ht]
    \vskip 0.2in
    \centering
    \begin{subfigure}[b]{1.0\textwidth}
        \centering
        \includegraphics[width=.32\linewidth]{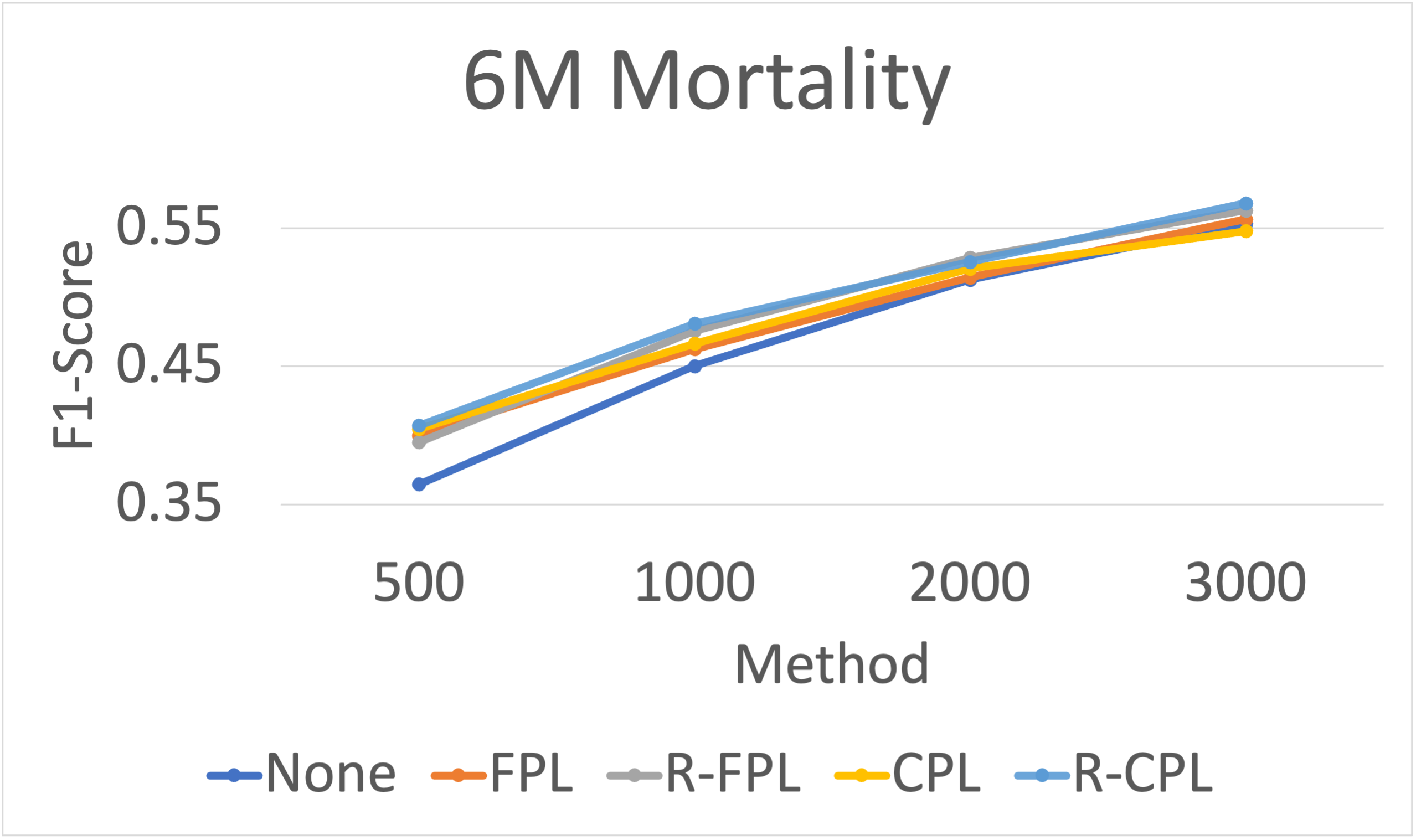}%
        \hfill
        \includegraphics[width=.32\linewidth]{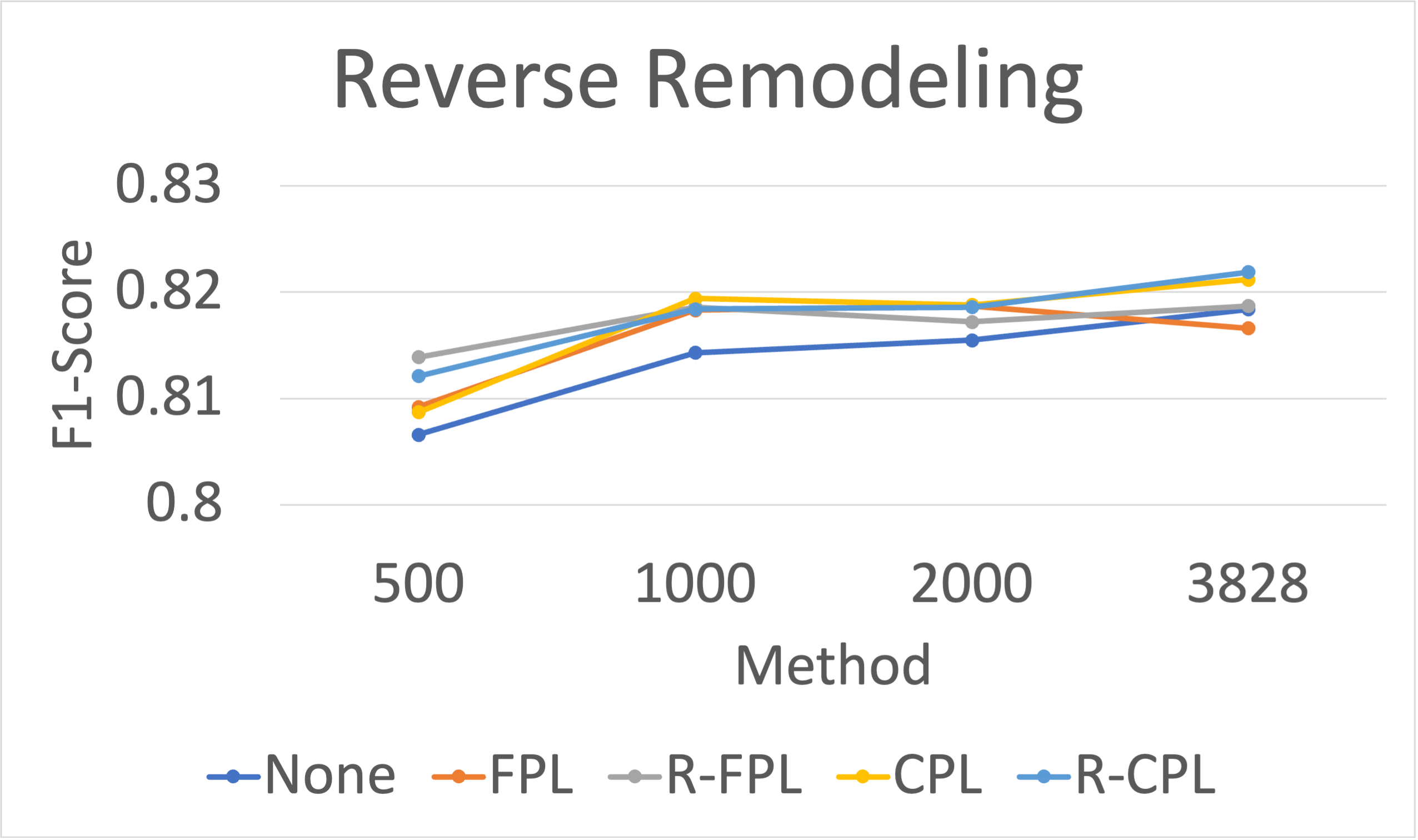}
        \hfill
        \includegraphics[width=.32\linewidth]{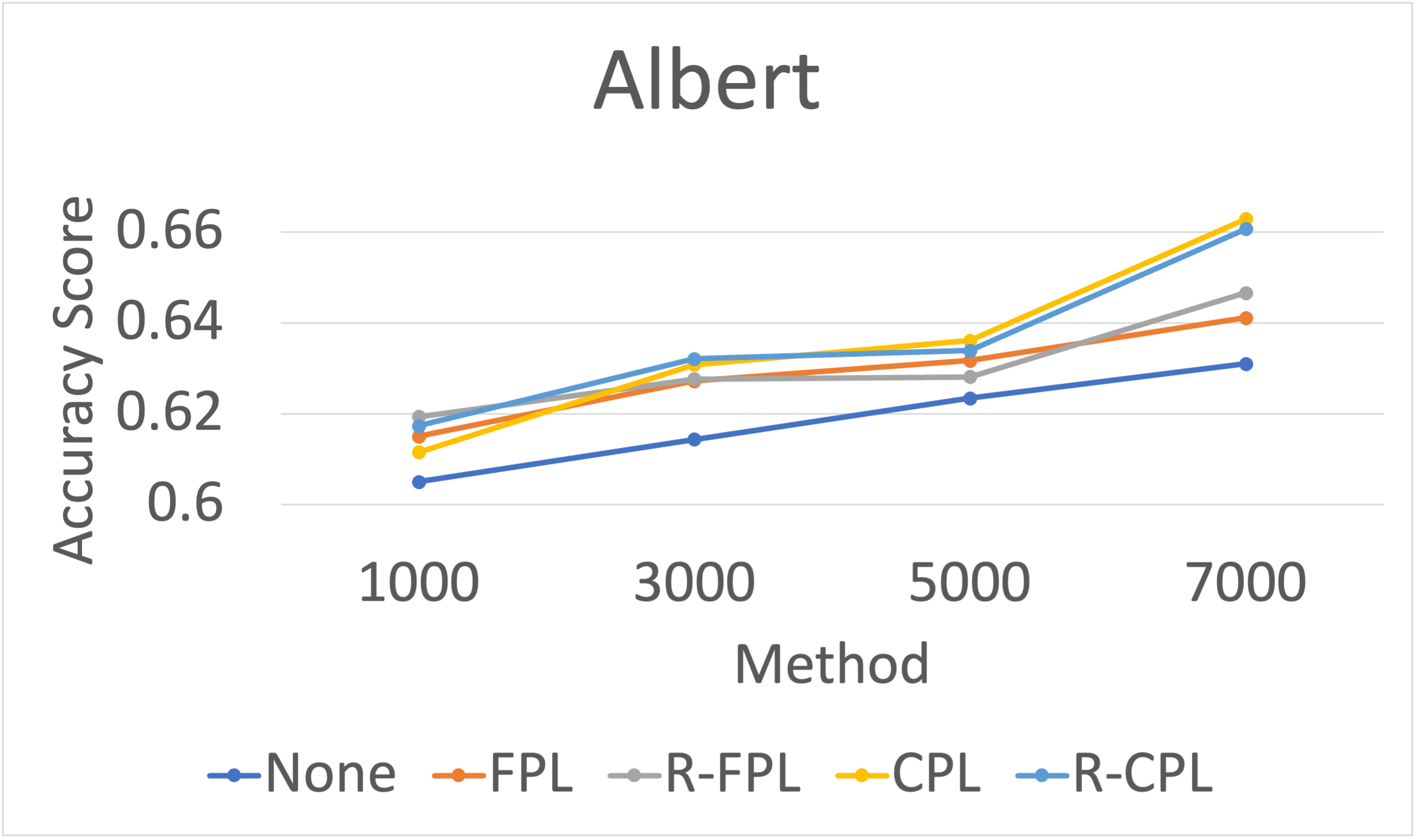}
        \caption{TabNet}
        \label{fig:fig2a-a}
    \end{subfigure} %\hfill
    \begin{subfigure}[b]{1.0\textwidth}
        \centering
        \includegraphics[width=.32\linewidth]{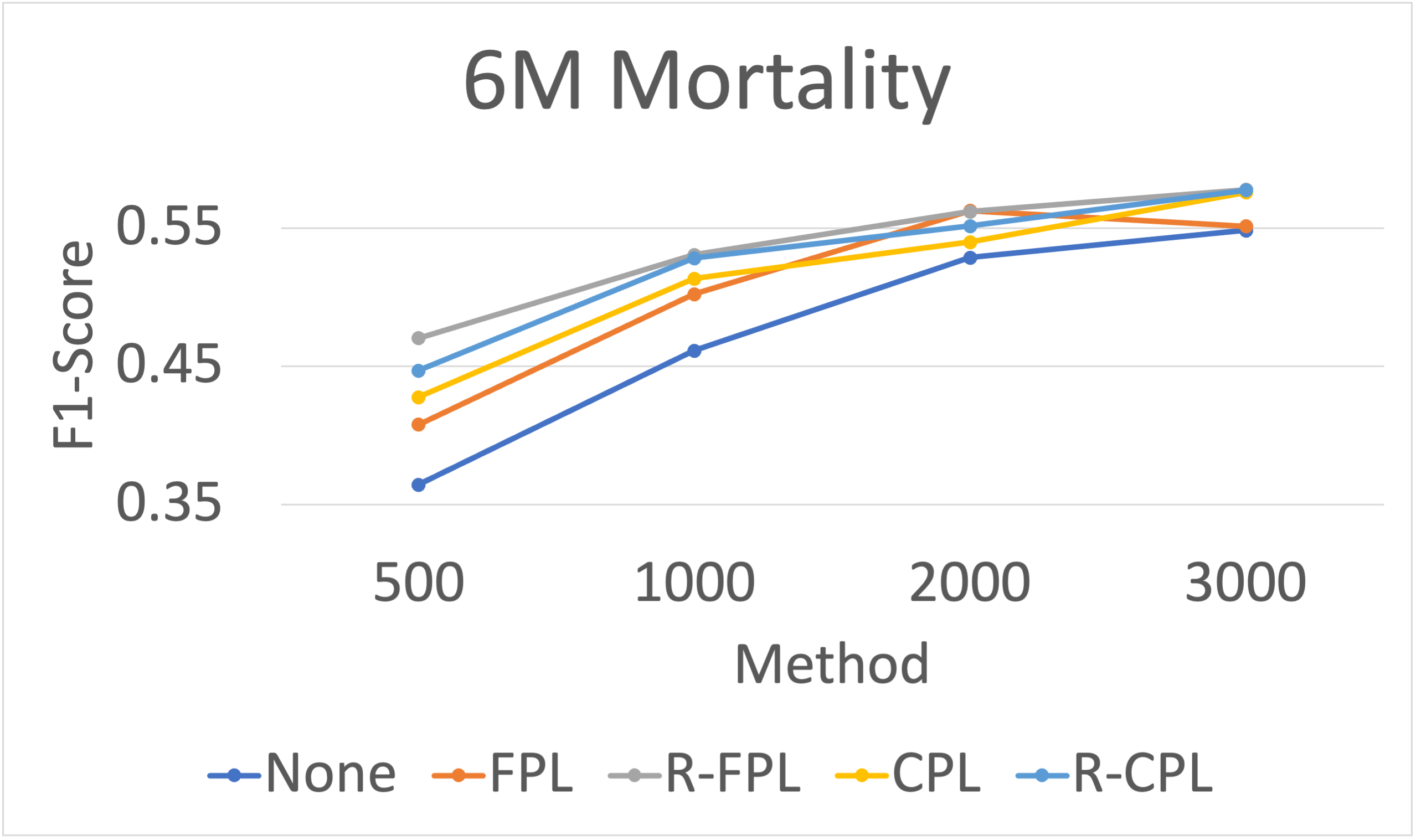}%
        \hfill
        \includegraphics[width=.32\linewidth]{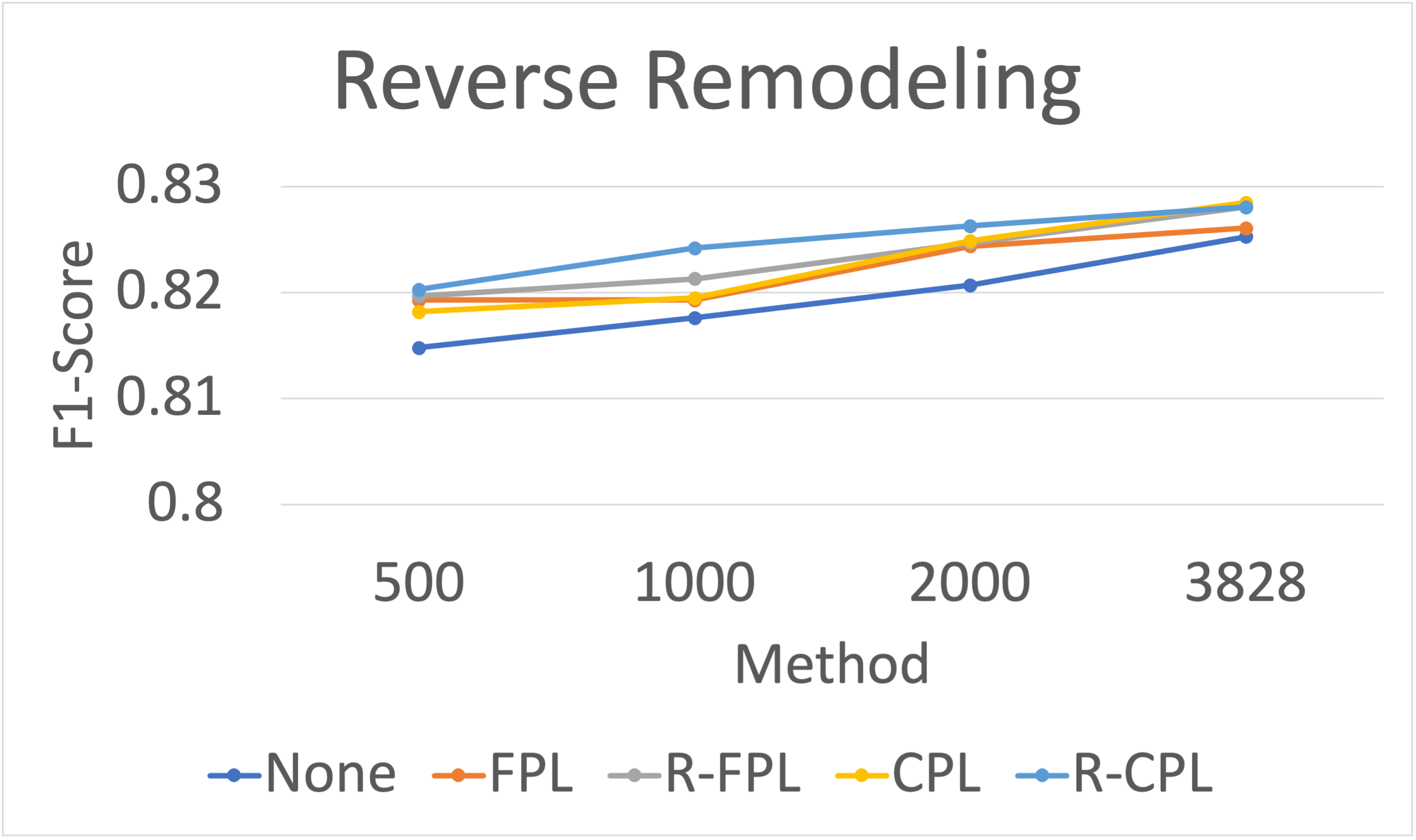}
        \hfill
        \includegraphics[width=.32\linewidth]{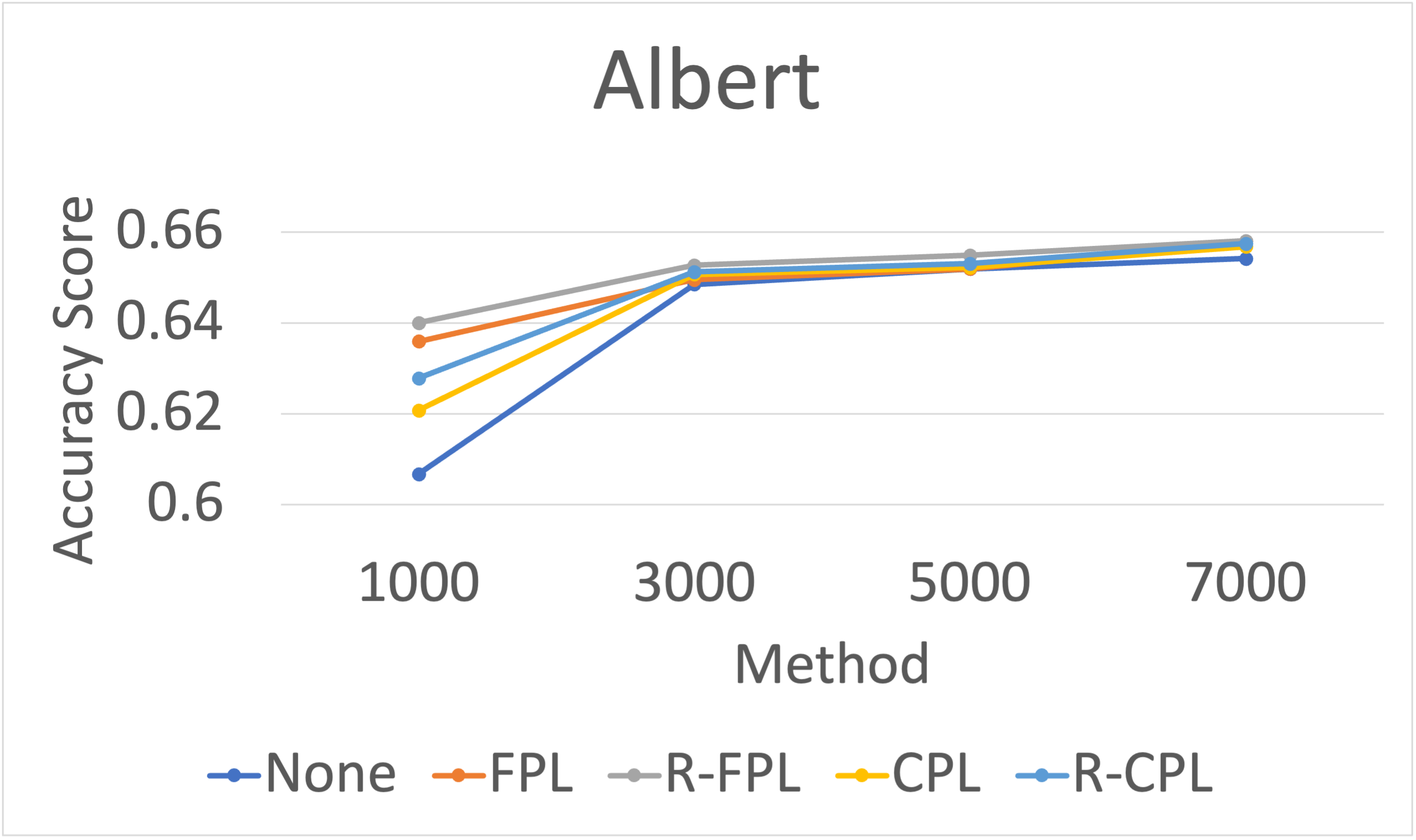}
        \caption{Saint}
        \label{fig:fig2a-b}
    \end{subfigure} %\hfill
    \caption{Performance comparison of one supervised learning without self-training method (NONE), and four pseudo-labeling methods: FPL, R-FPL, CPL, R-CPL using three datasets: 6 months mortality, 12 months reverse remodeling, and Albert with XGBoost and Saint, depending on the number of labeled data. As an evaluation metric for the 6 months mortality, and the reverse remodeling, we used F1-Score and, for Albert, accuracy score. (a) Performance comparison of the methods with XGBoost (b) Performance comparison of the methods with Saint.}
    \label{fig:fig2a}
    \vskip -0.2in
\end{figure*}
\FloatBarrier

\subsection{Table of Figure \ref{fig:fig2}}
\begin{table*}[h]
\caption{Performance comparison of one supervised learning without self-training method (NONE) and four pseudo-labeling methods: FPL, R-FPL, CPL, and R-CPL using three binary classification datasets: 6 months mortality, 12 months reverse remodeling, and Albert with pretrained TabNet, and pretrained Saint.
(a) F1-scores of the methods for the 6 months mortality dataset. (b) F1-scores of the methods for the 12 months reverse remodeling dataset. (c) Accuracy scores of the methods for click-through rate prediction for a given advertisement in Albert dataset. (d) The average of the ranks for each method.
}
\label{table:5}
\tiny
\vskip 0.15in
\begin{adjustwidth}{-0.5cm}{-1cm}
\begin{subtable}[h] {0.5\textwidth}
\begin{center}
\begin{sc}
\small
\caption{}
\label{table:6m_mortality_pre}
\begin{tabular}{lccr}
    \toprule
    Method & {\centering pretrained TabNet} & {\centering pretrained Saint} \\
    \midrule
    None & 0.3992 & 0.5447. \\ \hline
    FPL & 0.3992 & 0.5491 \\
    R-FPL & 0.3992 & 0.5598* \\ \hline
    CPL & 0.3992 & 0.5603  \\
    R-CPL & 0.3992 & \textbf{0.5796}* \\
    \bottomrule
\end{tabular}
\end{sc}
\end{center}
\end{subtable} %
\begin{subtable}[h] {0.5\textwidth}
\begin{center}
\begin{sc}
\small
\caption{}
\label{table:reverse_remodeling_pre}
\begin{tabular}{lccr}
    \toprule
    Method & {\centering pretrained TabNet} & {\centering pretrained Saint} \\
    \midrule
    None & 0.8116 & 0.8253 \\ \hline
    FPL & 0.8155* & 0.8262  \\
    R-FPL & 0.8151 & 0.8271*  \\ \hline
    CPL & 0.8124 & 0.8271  \\
    R-CPL & \textbf{0.8163}* & \textbf{0.8282}* \\
    \bottomrule
\end{tabular}
\end{sc}
\end{center}
\end{subtable} %
\vskip -0.1in
\vskip 0.15in
\begin{subtable}[h] {0.5\textwidth}
\begin{center}
\begin{sc}
\small
\caption{}
\label{table:albert_pre}
\begin{tabular}{lccr}
    \toprule
    Method & {\centering pretrained TabNet} & {\centering pretrained Saint} \\
    \midrule
    None & 0.6452 & 0.6444 \\ \hline
    FPL & 0.6502* & 0.6453 \\
    R-FPL & 0.6497 & 0.6467* \\ \hline
    CPL & 0.6513 & 0.6478 \\
    R-CPL & \textbf{0.6524}* & \textbf{0.6575}*  \\
    \bottomrule
\end{tabular}
\end{sc}
\end{center}
\end{subtable}%
\begin{subtable}[h] {0.5\textwidth}
\begin{center}
\begin{sc}
\small
\caption{}
\label{table:model_total_pre}
\begin{tabular}{lcc|cr}
    \toprule
    Method & {\centering pretrained TabNet} & {\centering pretrained Saint} & Avg\\
    \midrule
    None & 5.0 & 5.0 & 5.0 \\ \hline
    FPL & 2.5* & 4.0 & 3.3\\
    R-FPL & 3.5 & 2.7* & 3.1* \\ \hline
    CPL & 3.0 & 2.0 & 2.5 \\
    R-CPL & \textbf{1.0}* & \textbf{1.0}* & \textbf{1.0}* \\
    \bottomrule
\end{tabular}
\end{sc}
\end{center}
\end{subtable}%
\vskip -0.1in
\end{adjustwidth}
\end{table*}
\FloatBarrier

\section{Algorithm} \label{appendix:algorithm}
%\vskip -0.16in
\begin{algorithm}[H] 
   \caption{Self-Training Cycle}
   \label{alg:algorithm1}
\begin{algorithmic}
   \STATE {\bfseries Input:} Labeled Dataset $D_{L} = \{(x^{(i)}, y^{(i)})\}_{i = 1}^{N_{L}}$
   \STATE \quad\quad\quad Unlabeled Dataset $D_{U} = \{x^{(i)}, \varnothing\}_{i = 1}^{N_{U}}$
   \STATE \quad\quad\quad scoring function $f$ for input $x^{(i)}$
   \STATE \quad\quad\quad pseudo-labeler $\Phi_{l}(f, C, x^{(i)})$ where $C$ is a trained classifier
   \STATE \quad\quad\quad performance measure of a classifier $pm(C)$
   \newline
   \STATE $C_{new} \leftarrow$ trained classifiers with $D_{L}$
   \STATE $C_{B} \leftarrow C_{new}$
   \STATE $C_{old} \leftarrow$ a random initialized classifier
   
   % \WHILE{$pm(C_{new}) > pm(C_{old})$ or $|\Tilde{D}_{U}| \neq |{D}_{U}|$} 
   \WHILE{according to the of $\Phi_{l}$} 
   \STATE // $pm(C_{new}) > pm(C_{old})$ for fixed threshold $\Phi_{l}$
   \STATE // $|\Tilde{D}_{U}| \neq |{D}_{U}|$ for curriculum $\Phi_{l}$
   % \STATE $X_{U}$ $\leftarrow$ features of data from $D_{U}$ 
   \STATE $C_{old} \leftarrow C_{new}$
   \STATE $\Tilde{D}_{U}$ = $\varnothing$
   \FOR{$x^{(i)}$ in $X_U$}
        \STATE $\Tilde{D}_{U}$ = $\Tilde{D}_{U} \cup \Phi_{l}(f, C_{new}, x^{(i)})$
   \ENDFOR
   % \STATE $\Tilde{D}_{U}$ = $\Phi_{l}(f, C_{new}, X_{U})$
   \STATE $\Tilde{D} \leftarrow D_{L} \cup \Tilde{D}_{U}$

   \STATE $C_{new} \leftarrow$ a classifier newly trained with $\Tilde{D}$
   \IF{$pm(C_{new}) > pm(C_{B})$}
        \STATE $C_{B} \leftarrow C_{new}$
   \ENDIF
   %\FOR{$i=1$ {\bfseries to} $m-1$}
   %\ENDFOR
   \ENDWHILE
   \STATE {\bfseries Output:} $C_{B}$
\end{algorithmic}
\end{algorithm}

\section{Additional Experimental Results}
\vskip -0.2in
\begin{table}[H]
\caption{F1 score of Saint with one supervised learning without self-training method (NONE), four pseudo-labeling methods: FPL, R-FPL, CPL, R-CPL, and one pretraining method on the Bank Marketing dataset.}
\label{table:1a}
\vskip 0.15in
\begin{center}
\begin{scriptsize}
\begin{sc}
\begin{tabular}{lccccr}
\toprule
Method $\backslash$ N Labels & 500 & 1000 & 2000 & 3000 \\
\midrule
None         & 0.5066           & 0.5374           & 0.5472           & 0.5620 \\
FPL          & \textbf{0.5406}* & 0.5529           & 0.5630           & 0.5709 \\
R-FPL        & 0.5378           & \textbf{0.5601*} & 0.5661*          & 0.5780* \\
CPL          & 0.5328*          & 0.5513           & 0.5628           & 0.5749 \\
R-CPL        & 0.5282           & 0.5521*          & \textbf{0.5647}* & \textbf{0.5788}* \\ \hline
Pretraining  & 0.1499           & 0.5169           & 0.5416           & 0.5655 \\
\bottomrule
\end{tabular}
\end{sc}
\end{scriptsize}
\end{center}
\vskip -0.1in
\end{table}

\begin{table}[H]
\caption{Performance comparison of one supervised learning without self-training method (NONE), four pseudo-labeling methods: FPL, R-FPL, CPL, and R-CPL, and naive self-training \cite{pseudo_label} using ten binary classification datasets: 6 months mortality (6MM), 12 months reverse remodeling (12MRR), Albert, Christine, Jasmine, Madeline, Philippine, and Sylvine, Coupon, and Bank Marketing (Bank), and five multiclass classification datasets: Dilbert, Fabert, Splice, MNIST, Steel Plates Fault (Steel) with XGBoost. The performance metric used for 6MM, 12MRR, Bank, and Steel is F1-score, and for the rest accuracy score.}
\label{table:6}
\vskip 0.15in
\begin{center}
\begin{tiny}
\begin{sc}
\begin{adjustwidth}{-0.75cm}{-1cm}
\begin{tabular}{lp{0.04\textwidth}p{0.04\textwidth}p{0.04\textwidth}p{0.04\textwidth}p{0.04\textwidth}p{0.04\textwidth}p{0.04\textwidth}p{0.04\textwidth} p{0.04\textwidth}p{0.04\textwidth}p{0.04\textwidth}p{0.04\textwidth}p{0.04\textwidth}p{0.04\textwidth}p{0.04\textwidth}r}
\toprule
Method $\backslash$ Dataset & 6MM & 12MRR & Albert & Christine & Jasmine & Madeline & Philippine & Sylvine & Coupon & Bank & Dilbert & Fabert & Splice & MNIST & Steel \\
\midrule
None   & 0.5534 & 0.8184 & 0.6310 & 0.7071 & 0.7922 & 0.6930 & 0.7219 & 0.9290 & 0.7142 & 0.5635 & 0.9581 & 0.5872 & 0.9464 & 0.9639 & 0.7228 \\
FPL    & 0.5565 & 0.8166 & 0.6411 & \textbf{0.7148}* & 0.7932 & 0.7080 & 0.7305* & 0.9295* & 0.7143 & 0.5676 & 0.9639 & 0.5870 & 0.9529 & 0.9645 & 0.7269 \\
R-FPL  & 0.5630* & 0.8187* & 0.6466* & 0.7086 & 0.7989* & 0.7236* & 0.7274 & 0.9290 & 0.7215* & 0.5711* & \textbf{0.9646}* & \textbf{0.5890}* & \textbf{0.9545}* & \textbf{0.9647}* & 0.7300* \\
CPL    & 0.5479 & 0.8212 & \textbf{0.6629}* & 0.7111* & 0.7969 & 0.7080 & 0.7250 & 0.9297 & 0.7095 & 0.5707 & 0.9633 & 0.5875 & 0.9503 & 0.9644 & 0.7316 \\
R-CPL  & \textbf{0.5681}* & \textbf{0.8219}* & 0.6607 & 0.7082 & \textbf{0.7993}* & \textbf{0.7261}* & \textbf{0.7311}* & \textbf{0.9301}* & \textbf{0.7250}* & \textbf{0.5730}* & 0.9642* & 0.5881* & 0.9531* & 0.9646* & \textbf{0.7351}* \\ \hline
Naive  & 0.5613 & 0.8176 & 0.6583 & 0.7084 & 0.7956 & 0.7162 & 0.7299 & 0.9279 & 0.7046 & 0.5726 & 0.9639 & 0.5868 & 0.9513 & 0.9643 & 0.7316 \\
\bottomrule
\end{tabular}
\end{adjustwidth}
\end{sc}
\end{tiny}
\end{center}
\vskip -0.1in
\end{table}

\newpage

\section{Details of Datsets} \label{appendix_dataset}

\textbf{KAMIR}
Korea Acute Myocardial Infarction Registry (KAMIR) is a prospective and observational nation-wide multicenter registry investigating current management practices, risk factors, and clinical outcomes in Korean patients with acute myocardial infarction (AMI) beginning in 2003 sponsored by the Korean Society of Cardiology.
We obtained data of AMI patients between 2015 and 2020 to predict 6 months mortality and 12 months reverse remodeling (i.e.\ the restoration of heart function after AMI).
\begin{itemize}
\item{\textbf{6 months mortality} - Predicting 6 months mortality of each patient after AMI.}
\item{\textbf{12 months reverse remodeling} - Predicting 12 months reverse remodeling of each patient after AMI.}
% \item{\textbf{KAMIR} - 6 months mortality prediction and 12 months reverse remodeling prediction after AMI.} \\
\end{itemize}

\textbf{OpenML}
\begin{itemize}
\item{\textbf{Albert} - Predicting click-through rate after watching advertisements. 
This dataset was prepared from 'Criteos Delayed Feedback in Display Advertising' (CRITEO) dataset to use for AutoML Challeges.}

\item{\textbf{Christine} - Predicting chemical activity of molecules. 
This dataset was prepared from 'Activation of pyruvate kynase' data to use for AutoML Challenges.}

\item{\textbf{Jasmine} - Predicting cursive script subwords. 
This dataset was prepared from 'Arabic Manuscripts' data to use for AutoML Challenges.}

\item{\textbf{Madeline} - Predicting class of artificial continuous data which is very non-linear. 
This dataset was prepared from NIPS 2003 feature selection challenge to use for AutoML Challenges.}

\item{\textbf{Philippine} - Cell division (meiosis) classification.
This dataset was prepared from 'Mitosis classification' challenge to use for AutoML Challenges.}

\item{\textbf{sylvine} - Forest cover type (Krummholz or not) classification.
This dataset was prepared from 'Forest cover type' data to use for AutoML Challenges.}

\item{\textbf{Dilbert} - 3D object feature vector classification.
This dataset was prepared from 'NORB' dataset to use for AutoML Challenges.}

\item{\textbf{Fabert} - Human gesture recognition.
This dataset was prepared from 'LAP2014 Gesture Recognition Dataset using Skeleton features' dataset to use for AutoML Challenges.}

\item{\textbf{Splice} - Predicting DNA splice junctions.}
\item{\textbf{MNIST} - Predicting handwritten digits.}
\item{\textbf{Steel Plates Fault} - Predicting steel plates fault.}
\end{itemize}

More information of datasets which are processed for AutoML Challenges is at \url{https://automl.chalearn.org/home}.

\textbf{UCI Machine Learning Repository}

\begin{itemize}

\item{\textbf{Bank Marketing} - Predicting term deposit subscription in bank marketing.}
\item{\textbf{In-vehicle coupon recommendation} - Predicting recommended coupon acception.}

\end{itemize}

\begin{table}[h]
\caption{Additional details of datasets. 6 MM denotes the 6 months mortality dataset and 12 MRR denotes the 12 months reverse remodeling dataset. Note that, N labeled means the number of labeled samples to use in training, it is irrelevant to test data. The number of test data is \textbf{$\frac{1}{3}$ of total labeled samples.} Except reverse remodeling dataset which has total of 5,742 labels, N sample is equal to the number of total labels for all datasets.}
\label{sample-table}
\vskip 0.15in
\begin{center}
\begin{sc}
\begin{small}
\begin{tabular}{lccccccr}
\toprule
Data set & N sample & N labeled & N feature & N numerical feature & N class \\
\midrule
6 MM & 15,628 & 3000 & 76 & 29 & 2 \\
12 MRR & 15,628 & 3828 & 76 & 29 & 2 \\
Albert     & 20,000 & 7000 & 78 & 26 & 2 \\
Christine  & 5418 & 1000 & 1636 & 1636 & 2 \\
Jasmine    & 2984 & 600 & 144 & 136 & 2 \\
Madeline   & 3140 & 600 & 259 & 259 & 2 \\
Philippine & 5832 & 1000 & 308 & 308 & 2 \\
sylvine    & 5124 & 1000 & 20 & 2 & 2 \\
Dilbert    & 10000 & 3000 & 2000 & 2000 & 5 \\
Fabert     & 8237 & 1700 & 800 & 800 & 7 \\
Splice     & 3190 & 600 & 60 & 0 & 3 \\
MNIST      & 70000 & 15000 & 784 & 784 & 10 \\
Steel      & 1941 & 400 & 27 & 27 & 7 \\
Bank       & 45211 & 3000 & 16 & 6 & 2 \\
Coupon     & 12684 & 3000 & 23 & 1 & 2 \\
\bottomrule
\end{tabular}
\end{small}
\end{sc}
\end{center}
\vskip -0.1in
\end{table}

\newpage
%\vskip 10in

\section{Details of Hyperparameter Tunning} \label{appendix_hyperparameter}

\begin{table}[h]
\caption{Optuna hyperparameter search space for XGBoost}
\label{xgboost_search_space}
\vskip 0.15in
\begin{center}
\begin{sc}
\begin{tabular}{lcccr}
\toprule
Hyperparameter & Search Method & Search Space \\
\midrule
max\_leaves & suggest\_int & [300,4000] \\
n\_estimators & suggest\_int & [10,3000] \\
learning\_rate & suggest\_uniform & [0,1] \\
max\_depth & suggest\_int & [3, 20] \\
scale\_pos\_weight & suggest\_int & [1, 100] \\
\bottomrule
\end{tabular}
\end{sc}
\end{center}
\vskip -0.1in
\end{table}

\begin{table}[h]
\caption{Optuna hyperparameter search space for LightGBM}
\label{lightgbm_search_space}
\vskip 0.15in
\begin{center}
\begin{sc}
\begin{tabular}{lcccr}
\toprule
Hyperparameter & Search Method & Search Space \\
\midrule
max\_leaves & suggest\_int & [300,4000] \\
n\_estimators & suggest\_int & [10,3000] \\
learning\_rate & suggest\_uniform & [0,1] \\
num\_iterations & suggest\_int & [100,2000] \\
max\_depth & suggest\_int & [3, 50] \\
scale\_pos\_weight & suggest\_int & [1, 100] \\
\bottomrule
\end{tabular}
\end{sc}
\end{center}
\vskip -0.1in
\end{table}

\begin{table}[h]
\caption{Optuna hyperparameter search space for FT-Transformer}
\label{fttransformer_search_space}
\vskip 0.15in
\begin{center}
\begin{sc}
\begin{tabular}{lcccr}
\toprule
Hyperparameter & Search Method & Search Space \\
\midrule
input\_embed\_dim & suggest\_categorical & [16,24,32,48] \\
embedding\_dropout & suggest\_uniform & [0.05,0.3] \\
share\_embedding & suggest\_categorical & [True, False] \\
num\_heads & suggest\_categorical & [1,2,4,8] \\
num\_attn\_blocks & suggest\_int & [2,10] \\
transformer\_activation & suggest\_categorical & [GEGLU, ReGLU, SwiGLU] \\
use\_batch\_norm & suggest\_categorical & [True, False] \\
batch\_norm\_continuous\_input & suggest\_categorical & [True, False] \\
learning\_rate & suggest\_uniform & [0.0001, 0.05] \\
scheduler\_gamma & suggest\_uniform & [0.1, 0.95] \\
scheduler\_step\_size & suggest\_int & [10, 100] \\ 
\bottomrule
\end{tabular}
\end{sc}
\end{center}
\vskip -0.1in
\end{table}

\begin{table}[h]
\caption{Optuna hyperparameter search space for TabTransformer}
\label{tabtransformer_search_space}
\vskip 0.15in
\begin{center}
\begin{sc}
\begin{tabular}{lcccr}
\toprule
Hyperparameter & Search Method & Search Space \\
\midrule
input\_embed\_dim & suggest\_categorical & [16,24,32,48] \\
embedding\_dropout & suggest\_uniform & [0.05,0.3] \\
share\_embedding & suggest\_categorical & [True, False] \\
num\_heads & suggest\_categorical & [1,2,4,8] \\
num\_attn\_blocks & suggest\_int & [2,10] \\
transformer\_activation & suggest\_categorical & [GEGLU, ReGLU, SwiGLU] \\
use\_batch\_norm & suggest\_categorical & [True, False] \\
batch\_norm\_continuous\_input & suggest\_categorical & [True, False] \\
learning\_rate & suggest\_uniform & [0.0001, 0.05] \\
scheduler\_gamma & suggest\_uniform & [0.1, 0.95] \\
scheduler\_step\_size & suggest\_int & [10, 100] \\ 
\bottomrule
\end{tabular}
\end{sc}
\end{center}
\vskip -0.1in
\end{table}

\begin{table}[h]
\caption{Optuna hyperparameter search space for Saint}
\label{saint_search_space}
\vskip 0.15in
\begin{center}
\begin{sc}
\begin{tabular}{lcccr}
\toprule
Hyperparameter & Search Method & Search Space \\
\midrule
transformer\_depth & suggest\_int & [1, 6] \\
attention\_heads & suggest\_int & [1, 6] \\
attention\_dropout & suggest\_uniform & [0.05, 0.3] \\
ff\_dropout & suggest\_uniform & [0.05, 0.3] \\
lr & suggest\_uniform & [0.00001, 0.01] \\
embedding\_size & suggest\_categorical & [4, 12, 16, 24, 32, 48] \\
attentiontype & suggest\_categorical & [col, colrow, row] \\
optimiezr & suggest\_categorical & [AdamW, Adam, SGD] \\
scheduler & suggest\_categorical & [cosine, linear] \\
epochs & suggest\_categorical & [50, 100, 150, 200, 250] \\
batchsize & suggest\_categorical & [128, 256, 512] \\
pretrain\_epochs & suggest\_categorical & [50, 100] \\
pt\_aug & suggest\_categorical & [mixup, cutmix, [mixup, cutmix]] \\
\bottomrule
\end{tabular}
\end{sc}
\end{center}
\vskip -0.1in
\end{table}

\begin{table}
\caption{Optuna hyperparameter search space for MLP}
\label{mlp_search_space}
\vskip 0.15in
\begin{center}
\begin{sc}
\begin{adjustwidth}{-0.75cm}{-1cm}
\begin{tabular}{lcccr}
\toprule
Hyperparameter & Search Method & Search Space \\
\midrule
embedding\_dropout & suggest\_uniform & [0, 0.2] \\
layers & suggest\_categorical & [128-64-32, 256-128-64, 128-64-32-16, 256-128-64-32] \\
activation & suggest\_categorical & [ReLU, LeakyReLU] \\
use\_batch\_norm & suggest\_categorical & [True, False] \\
batch\_norm\_continuous\_input & suggest\_categorical & [True, False] \\
learning\_rate & suggest\_uniform & [0.0001, 0.05] \\
scheduler\_gamma & suggest\_uniform & [0.1, 0.95] \\
scheduler\_step\_size & suggest\_int & [10, 100] \\
\bottomrule
\end{tabular}
\end{adjustwidth}
\end{sc}
\end{center}
\vskip -0.1in
\end{table}

\vspace{0.1cm}

\begin{table}[h]
\caption{Optuna hyperparameter search space for TabNet}
\label{ptabnet_search_space}
\vskip 0.15in
\begin{center}
\begin{sc}
\begin{tabular}{lcccr}
\toprule
Hyperparameter & Search Space \\
\midrule
n\_d & suggest\_int & [8, 32] \\
n\_steps & suggest\_int & [3, 10] \\
n\_independent & suggest\_int & [1, 5] \\
n\_shared & suggest\_int & [1, 5] \\
lr & suggest\_uniform & [0.0001, 0.05] \\
gamma & suggest\_uniform & [1.0, 2.0] \\
mask\_type & suggest\_categorical & [entmax, sparsemax] \\
cat\_emb\_dim & suggest\_categorical & [1, 2, 4, 8] \\
lambda\_sparse & suggest\_uniform & [3e-5, 3e-3] \\
scheduler\_fn & suggest\_categorical & [None, StepLR, CosineAnnealingLR] \\
pretraining\_ratio & suggest\_uniform & [1e-5, 1] \\
\bottomrule
\end{tabular}
\end{sc}
\end{center}
\vskip -0.1in
\end{table}

\end{document}